\pdfoutput=1
\PassOptionsToPackage{dvipsnames}{xcolor}
\documentclass[11pt]{article}

\usepackage{EMNLP2023}

\usepackage{times}
\usepackage{latexsym}

\usepackage[T1]{fontenc}
\usepackage[utf8]{inputenc}

\usepackage{microtype}

\usepackage{inconsolata}

\usepackage{hyperref}       % hyperlinks
\usepackage{url}            % simple URL typesetting
\usepackage{booktabs}       % professional-quality tables
\usepackage{amsfonts}       % blackboard math symbols
\usepackage{nicefrac}       % compact symbols for 1/2, etc.
\usepackage{xcolor,colortbl}         % colors
\usepackage{lipsum}
\usepackage{caption}
\usepackage{subcaption}
\usepackage{soul}
\usepackage{graphicx}
\usepackage[multiple]{footmisc}
\usepackage{multirow}
\usepackage{makecell}
\usepackage{enumitem}

\definecolor{non_np_color}{HTML}{2596BE}
\definecolor{np_color}{HTML}{E33E75}

\newcommand{\negp}[0]{$\mathrm{NegP}$}
\newcommand{\negpbold}[0]{\textcolor{np_color}{$\mathbf{NegP}$}}
\newcommand{\others}[0]{$\mathrm{Others}$}
\newcommand{\othersbold}[0]{\textcolor{non_np_color}{$\mathbf{Others}$}}

\setlist{topsep=3pt,itemsep=3pt,partopsep=3pt, parsep=3pt}

\title{Negative Object Presence Evaluation (NOPE) to \\ Measure Object Hallucination in Vision-Language Models}
 \author{Holy Lovenia\thanks{\hspace{0.2cm}The majority of the work was done when the author was studying at HKUST.}\hspace{0.15cm}$^{,1,2}$ \quad Wenliang Dai\thanks{\hspace{0.2cm}Joint second authors.}\hspace{0.15cm}$^{,1}$ \quad Samuel Cahyawijaya$^{\dagger,1}$ \quad Ziwei Ji$^{1}$ \quad Pascale Fung$^{1}$ \\
  $^{1}$ The Hong Kong University of Science and Technology \\
  $^{2}$ AI Singapore \\
  \texttt{holy@aisingapore.org, pascale@ust.hk} \\}

\begin{document}
\maketitle

\begin{abstract}
Object hallucination poses a significant challenge in vision-language (VL) models, often leading to the generation of nonsensical or unfaithful responses with non-existent objects.
However, the absence of a general measurement for evaluating object hallucination in VL models has hindered our understanding and ability to mitigate this issue.
In this work, we present NOPE (Negative Object Presence Evaluation), a novel benchmark designed to assess object hallucination in VL models through visual question answering (VQA). We propose a cost-effective and scalable approach utilizing large language models to generate 29.5k synthetic negative pronoun (\negp) data of high quality for NOPE.
We extensively investigate the performance of 10 state-of-the-art VL models in discerning the non-existence of objects in visual questions, where the ground truth answers are denoted as \negp~(e.g., "none"). Additionally, we evaluate their standard performance on visual questions on 9 other VQA datasets.
Through our experiments, we demonstrate that no VL model is immune to the vulnerability of object hallucination, as all models achieve accuracy below 10\% on \negp. Furthermore, we uncover that lexically diverse visual questions, question types with large scopes, and scene-relevant objects capitalize the risk of object hallucination in VL models.
\end{abstract}

\section{Introduction}
\label{sec:intro}

% Generative capabilities of deep learning based models have advanced in tremendous leaps over the recent years for both unimodal (e.g., natural language processing and computer vision) and multimodal tasks.

% VQA has been the go-to task to measure a model's skills in understanding natural language and visual inputs as well as reasoning over the knowledge obtained from these different modalities~\cite{antol2015vqa}. In visual question answering (VQA), given an image and a natural language question about the image, a model is expected to infer and then generate an accurate natural language answer.
% Over the recent years, there has been an abundance of studies working on various methods, models, and/or learning strategies to close the gap between human performance and model performance~\cite{yang2021tap, yi2018neural, zhou2020unified, ray2019sunny, gokhale2020mutant}, as well as studies constructing more rigorous and challenging VQA benchmarks to raise the performance standard even higher~\cite{antol2015vqa, sheng2021human, li2021adversarial, goyal2017making, marino2019ok}.

\begin{figure}[t]
  \centering
  \includegraphics[width=\linewidth]{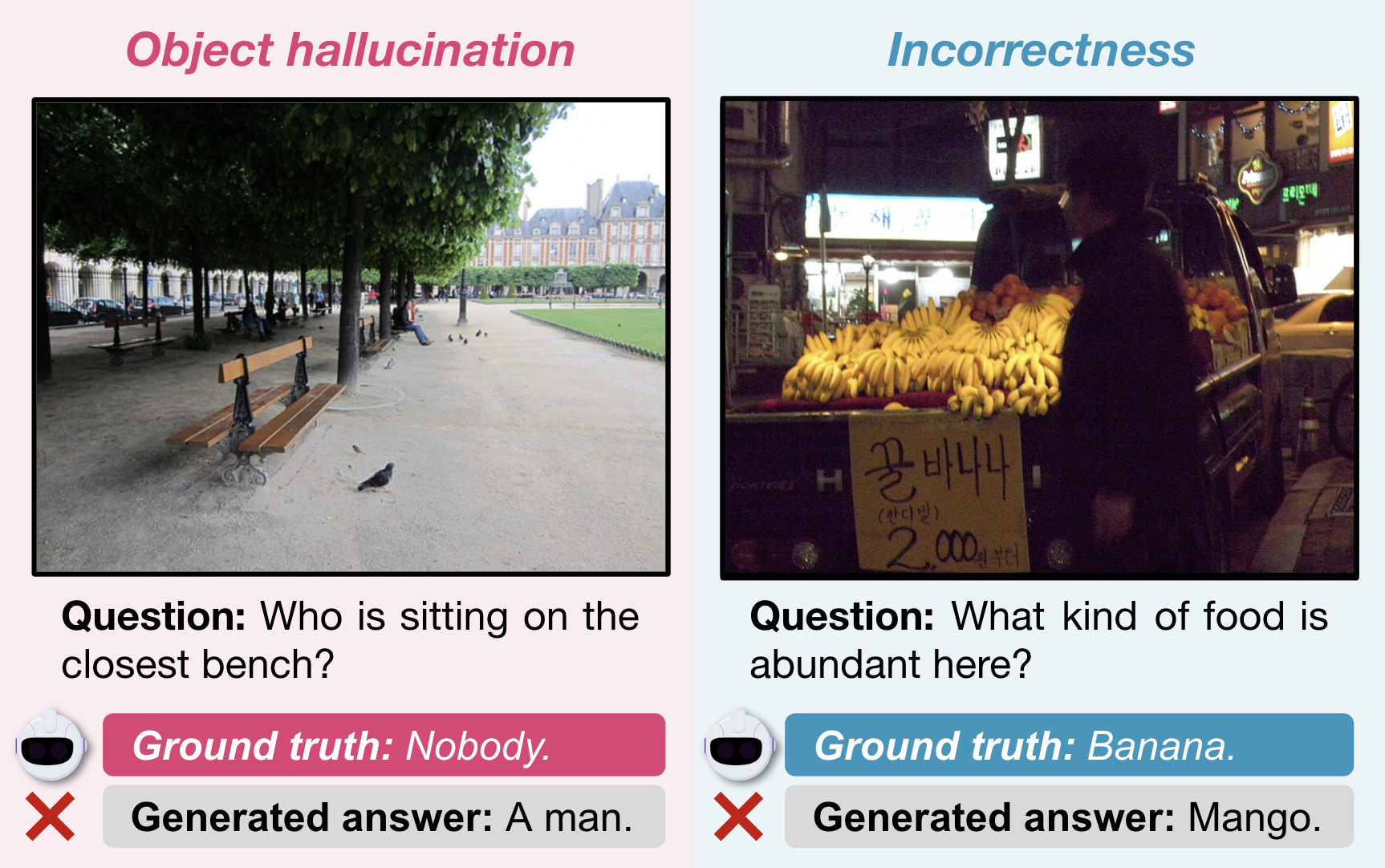}
  \caption{Example of object hallucination and incorrectness in VQA. The model hallucinates a non-existent man sitting on the closest bench in the left image, while in the right image, it simply answers inaccurately.}
  \label{fig:objhall-vs-incorrectness}
  \vspace{-7pt}
\end{figure}

% Over the recent years, there has been an abundance of vision-language (VL) studies working on various methods, models, and learning strategies to close the gap between human performance and model performance~\cite{yang2021tap, yi2018neural, zhou2020unified, ray2019sunny, gokhale2020mutant, dai2021multimodal, vlkd, ishii2021erica, lovenia2022every, ji2022vscript, lovenia2023one}, as well as studies constructing more rigorous and challenging VL benchmarks to raise the performance standard even higher~\cite{antol2015vqa, sheng2021human, li2021adversarial, goyal2017making, marino2019ok}.
% Despite these efforts, VL models still suffer from the object hallucination problem, with generated responses containing non-existent objects in the input images
% that are nonsensical or unfaithful to the provided inputs by hallucinating non-existent objects
% ~\cite{ji2022survey,rohrbach2018object,dai2022plausible,kayhan2021hallu-od}.
In recent years, vision-language (VL) research has witnessed a proliferation of studies focusing on diverse methods, models, and learning strategies aimed at bridging the performance gap between human and model capabilities~\cite{yang2021tap, yi2018neural, zhou2020unified, ray2019sunny, gokhale2020mutant, dai2021multimodal, vlkd, ishii2021erica, lovenia2022every, ji2022vscript, lovenia2023one}. Furthermore, researchers have constructed more rigorous VL benchmarks to continually raise the performance standard~\cite{antol2015vqa, sheng2021human, li2021adversarial, goyal2017making, marino2019ok}.
However, despite these efforts, VL models continue to grapple with the persistent issue of object hallucination, where generated responses unfaithfully contain objects non-existent in the input images~\cite{ji2022survey,rohrbach2018object,dai2022plausible,kayhan2021hallu-od}.
As illustrated in Figure~\ref{fig:objhall-vs-incorrectness}, the failure of the model to faithfully ground the visual input leads to the production of unfaithful answers. These instances of object hallucination not only result in incorrect responses but also shed light on fundamental issues within VL models, such as over-reliance on unimodal priors~\cite{jing2020overcoming, agrawal2018don, gupta2022swapmix, niu2021counterfactual} and statistical bias~\cite{agrawal2016analyzing, goyal2017making, agarwal2020towards}. These underlying problems impede the models' ability to comprehend the concept of non-existence.

Despite the critical importance of addressing object hallucination in VL models, only a limited number of previous works have focused on mitigating this issue, primarily due to the challenges posed by the existing evaluation method in terms of generalization and scalability. CHAIR~\cite{rohrbach2018object} has primarily concentrated on evaluating non-existent objects based on handcrafted parsing criteria as well as a predefined list of object categories and their synonyms in the context of image captioning tasks, typically utilizing 80 object categories from MSCOCO~\cite{rohrbach2018object, biten2022let, yi2018neural}. However, the applicability of CHAIR to other datasets requires the generation of a new object category list, which exhibits varying levels of granularity across different studies~\cite{dai2022plausible,biten2022let}.

In this paper, we present NOPE (\textbf{N}egative \textbf{O}bject \textbf{P}resence \textbf{E}valuation) to quantitatively assess object hallucination through VQA. We establish a clear distinction between object hallucination and incorrectness as follows: a) \textbf{object hallucination} refers to the phenomenon in VQA where a VL model's response includes a non-existent object, despite the ground truth answer being a negative indefinite pronoun (e.g., "none", "no one", "nobody", "nowhere", "neither")~\cite{quirk1985negativepronoun} (\negpbold); and b) \textbf{incorrectness} occurs when a VL model fails to accurately respond to a question with a ground truth answer that is anything other than \negp, denoted as \othersbold~= $\mathbb{P} \setminus$\negp, where $\mathbb{P}$ represents the set of all phrases. By leveraging \negp, we evaluate object hallucination in NOPE, while \others~allows us to assess normative correctness across diverse corpora. Our contributions are as follows:

% In this work, we introduce NOPE (\textbf{N}egative \textbf{O}bject \textbf{P}resence \textbf{E}valuation) to assess object hallucination through VQA as a quantifiable measure. To separate the notion of object hallucination and incorrectness, we define that: a) \textbf{object hallucination} in VQA is a phenomenon in which a VL model answers a question whose ground truth answer is a negative indefinite pronoun (i.e., "none", "no one", "nobody", "nowhere", "neither")~\cite{quirk1985negativepronoun} with a non-existent object as if the object was present in the visual input; and b) \textbf{incorrectness} is when a VL model that fails to answer a question whose ground truth answer is anything but a negative indefinite pronoun, i.e., \others~$=\mathbb{P}\setminus$\negp, where $\mathbb{P}$ is a set of all phrases. Hence, we build NOPE using negative pronoun (\negp) VQA pairs to evaluate the rate of object hallucination and non-negative pronoun  (\others) VQA pairs to evaluate the normative correctness from various corpora.

\begin{enumerate}
    \item By utilizing NOPE, we construct a VQA diagnostic benchmark to measure the object hallucination rate of VL models. Our experiment covers a balanced proportion of \negp~and \others~data with a total of $\sim$30k and $\sim$36k data in the dev and test sets, and includes 10 state-of-the-art VL baselines performances.
    % \item On top of accuracy, we also introduce a simple model-based metric for \negp~answer assessment on the benchmark.
    We provide an in-depth analysis of the performances and limitations of the baselines. % and contrast them against human performance.
    \item We propose a novel automatic data generation pipeline to produce high-quality \negp~VQA data from existing image captioning data by multi-turn prompting instruction-tuned large language models (LLMs). We verify and analyze our generated \negp~data through automatic validation and human validation. Our \textbf{list-then-rewrite} method produces high-quality \negp~VQA data with 92\% validity.
    % \item Through NOPE, we present an in-depth analysis to answer the question ``\textcolor{red}{What factors capitalize the risk of object hallucination in vision-language models?}''. which serves as the foundation for mitigating object hallucination in VL models.
    \item Through extensive analysis in NOPE, we find that VL models tend to hallucinate more on data with higher lexical diversity, more scene-relevant objects, and larger answer scopes.
\end{enumerate}

\section{Related Work}
\label{sec:related-work}

\subsection{Hallucination in Vision-Language}

% The vast majority of works studying hallucination in vision-language 

Only a few works study hallucination in vision-language, with the vast majority of them focusing on the task of image captioning.
% \paragraph{Image captioning}
~\citet{rohrbach2018object} propose CHAIR, an automatic evaluation metric to measure object hallucination in generated image captions, which is defined as a phenomenon where the models produce captions containing objects that do not exist in the input visual context.~\citet{rohrbach2018object, dai2022plausible, sharma2018conceptual} also show that standard captioning metrics, e.g., CIDEr~\cite{vedantam2015cider}, METEOR~\cite{banerjee2005meteor}, SPICE~\cite{niu2022spice}, under-penalize object hallucination.
% ~\citet{wang2022measuring} find that 65\%-80\% of the incorrect system-generated captions are hallucinations.
These evaluations open up a way for efforts to mitigate hallucination in image captioning~\cite{biten2022let, zhang2021consensus, xiao2021hallucination, dai2022plausible}.
% These evaluations open up a way for efforts to mitigate hallucination in image captioning through co-occurrence matrix uniformization~\cite{biten2022let}, consensus graph representation learning framework~\cite{zhang2021consensus}, uncertainty-aware beam search (UABS)~\cite{xiao2021hallucination}, and object-masked language modeling pre-training objective~\cite{dai2022plausible}.
Concurrent to our work, \citet{li2023evaluating} propose POPE and frame the task of evaluating object hallucination as a binary-class VQA with only "yes/no" answer.
% No work has ever attempted to evaluate object hallucination in VQA.

% \subsection{Evaluating Hallucination}
% \dummy{\lipsum[1]}

% \subsection{\negpbold~and \othersbold~in Existing VQA Data}
% \label{sec:np-vs-non-np}

% \begin{figure}[b]
%   \centering
%   \includegraphics[width=1.0\linewidth]{images/vqa-h-np-vs-non-np-vqa.png}
%   \caption{The proportions of \negp~and \others~data in 10 existing VQA corpora.}
%   \label{fig:np-vs-non-np}
% \end{figure}

% Figure~\ref{fig:np-vs-non-np} illustrates the proportions of \negp~and \others~in 10 existing VQA datasets. Only $\sim$0.5\% of the existing VQA datasets are \negp, which are not sufficient to assess object hallucination with, so we present our \negp~data generation in \S\ref{sec:np-data-generation}.

\subsection{Question Generation for VQA Data}

% Various motivations inspire VQA data constructions. %, whether they are human-generated, machine-generated, or hybrid.
Most works rely on human annotators to generate visual questions with ensured quality: VQAv2.0 and VQAv1.0~\cite{goyal2017making, antol2015vqa}, Visual Genome~\cite{krishnavisualgenome}, Visual7W~\cite{zhu2016visual7w}, AdVQA~\cite{sheng2021human}, Vizwiz~\cite{gurari2018vizwiz, gurari2019vizwiz}, TextVQA~\cite{Singh_2019_CVPR}, R-VQA~\cite{lu2018rvqa}, VQA-Rephrasings~\cite{shah2019cycle}, etc.
% Most works rely on human annotators to generate visual questions with ensured quality: VQAv2.0 and VQAv1.0~\cite{goyal2017making, antol2015vqa}, Visual Genome~\cite{krishnavisualgenome}, Visual7W~\cite{zhu2016visual7w} that asserts a semantic link between textual descriptions and local image regions by object-level grounding, AdVQA~\cite{sheng2021human} that utilizes human adversarial to generate tricky questions that usually models fail to answer, Vizwiz~\cite{gurari2018vizwiz, gurari2019vizwiz} and TextVQA~\cite{Singh_2019_CVPR} that cater to the needs of visually impaired users, R-VQA~\cite{lu2018rvqa} that provides visual relation facts for semantic reasoning, VQA-Rephrasings~\cite{shah2019cycle} that focuses on self-consistency, etc.

However, the cost of human annotation is expensive, thus encouraging the exploration of a more scalable option: automatic VQA data generation.
~\citet{ren2015exploring} present a simple question generation algorithm with a syntactic parser to convert image descriptions into QA forms.~\citet{johnson2017clevr} use a functional program to generate synthetic images of objects as well as their relationships and relevant QA pairs using the ground-truth annotations.~\citet{kafle2017analysis} populate multiple question templates with the image annotations (e.g., region descriptions, relationship graphs, bounding boxes) obtained from image captioning data to construct TDIUC.~\citet{changpinyo2022all} annotate candidate answers by syntactically parsing the captions, then derive questions from them.
% using neural models for question generation and QA verification. 
While prior studies focus on generating \others~VQA data, we aim to generate \negp~VQA data, which has never been done by past works.
% While prior studies focus on generating \others~VQA data, our work aims to generate \negp~VQA data, which has never been done by past works.

% \subsection{Data Generation Using LLMs}

% Recent works have demonstrated that LLMs are effective at following natural
% language instructions after instruction tuning~\cite{mishra2022cross, wei2022finetuned, sanhmultitask, ouyang2022training, muennighoff2022crosslingual, bang2023multitask, iyer2022opt, touvron2023llama, du2022glm, zeng2023glm-130b}. The prompting abilities of LLMs have been utilized for automatic data generation in various tasks: instructions~\cite{wang2022self}, code-mixing~\cite{yong2023prompting}, text classification~\cite{wang2021towards}, natural language inference (NLI)~\cite{liu-etal-2022-wanli}, semantic textual similarity~\cite{schick-schutze-2021-generating}, educational question generation~\cite{wang2022towards}, and more.

\section{NOPE to Overcome Limited \negpbold}
% \section{\negpbold~VQA Data Generation}
\label{sec:np-vs-non-np}
\label{sec:np-data-generation}

\begin{figure}[t]
  \centering
  \includegraphics[width=0.85\linewidth]{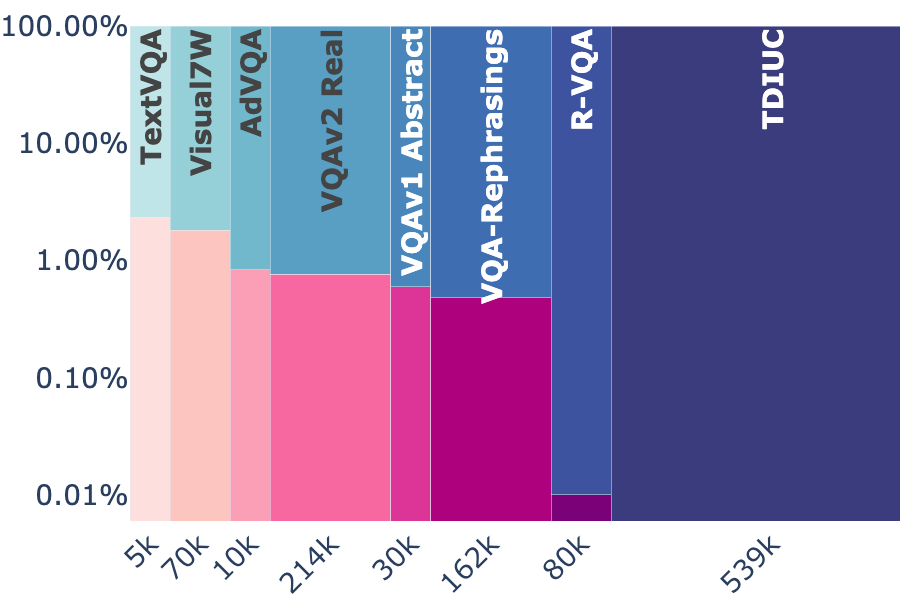}
  \caption{Only 0.4\% of existing VQA corpora consist of \textcolor{np_color}{\negp~}data. The rest 99.6\% is \textcolor{non_np_color}{\others}.}
  \label{fig:np-vs-non-np}
  \vspace{-5pt}
\end{figure}

As shown in Figure~\ref{fig:np-vs-non-np}, there is only a minuscule amount of \negp~data in the existing VQA datasets. In total, there are only $\sim$0.4\% of the existing VQA datasets are \negp, which are not sufficient to assess object hallucination in VL. For this reason, we create NOPE through a novel \negp~data generation method that aims to produce questions whose ground truth answers point to the absence of appropriate existent objects. Such ground truth \negp~answers are denoted as $A^\mathrm{NegP} = \{"none", "nothing", "nowhere", "zero", "0", \\ "no\;one", "nobody", "neither"\}$. We automatically generate synthetic \negp~VQA data by leveraging the zero-shot prompting abilities of pre-trained LLMs. To ensure the quality, we analyze the generated synthetic \negp~VQA data through both automatic and manual human evaluation. The resulting \negp~dataset is referred to as NOPE (\textbf{N}egative \textbf{O}bject \textbf{P}resence \textbf{E}valuation).

\subsection{Prompting Methodology}
\label{sec:prompt-method}

We utilize an image captioning dataset $\mathcal{D}_{cap} = \{(v_i, c_i, l_i\}^n_{i=1}$, where $v_i$ denotes a visual context, $c_i$ denotes a textual caption, and $l_i$ denotes the relevant image label annotations (i.e., names of objects in $v_i$). We rely on $c_i$ to describe the objects and the relationship between objects depicted in $v_i$. We explore two prompting methods with varying degrees of flexibility to generate \negp~questions from image captions: \textbf{generate-from-scratch} and \textbf{list-then-rewrite}. For clarity, we include all prompt templates with the examples in Appendix~\ref{sec:prompt-templates} and the automatic validation methods to ensure the validity of the generated questions in Appendix~\ref{sec:auto-validation-methods}.

\paragraph{Generate-from-scratch}

In this method, we prompt an LLM to generate a question $q_i$ given three different variables: 1) an interrogative word $w_i \in \{"what", "where", "how\;many", "who", \\ "which"\}$ to assert the question type needed for $q_i$, 2) a ground truth \negp~answer $a_i \in A^\mathrm{NegP}$ that matches $w_i$, and 3) an image caption $c_i$.

\begin{figure}[!t]
  \centering
  \includegraphics[width=0.85\linewidth]{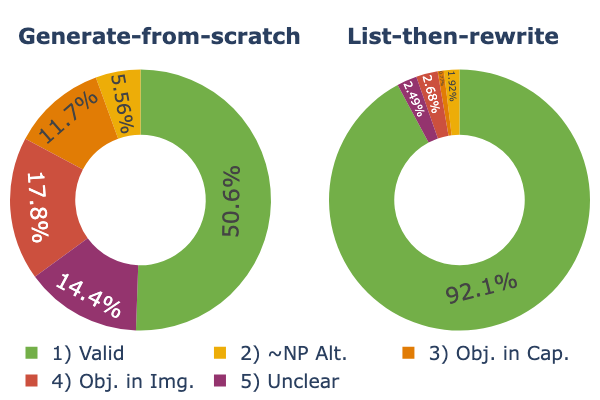}
  \caption{Human evaluation results of \textcolor{np_color}{\negp} questions by \textbf{generate-from-scratch} and \textbf{list-then-rewrite} according to the categories in \S\ref{sec:human-eval-guidelines}.}
  \label{fig:simple-human-eval}
  \vspace{-5pt}
\end{figure}

\paragraph{List-then-rewrite}

LLMs can infer conversational contexts and follow instructions over multiple turns~\cite{nijkamp2023codegen, volum-etal-2022-craft, bang2023multitask}.
% Recent works show that LLMs have the capacity to understand long inputs and generate coherent responses, enabling them to infer conversational contexts and follow instructions over multiple turns~\cite{nijkamp2023codegen, volum-etal-2022-craft, bang2023multitask}. 
Leveraging this multi-turn capability of LLMs, we frame our question generation task into two steps. (1) For object listing, given an image caption $c_i$ and the relevant object annotations $l_i$, we prompt an LLM to list $m$ objects $o_i = \{o_{i,j}\}^m_{j=1}$ that are ``closely related''\footnote{\label{fn:diff-level}We use ``closely related'' (hard) for brevity. However, this object-scene relevance can be switched to ``loosely related'' or ``completely unrelated'' in practice.} but not mentioned. (2) For question rewriting, the LLM has to paraphrase a provided reference question, which is sourced from a diverse pool of human-generated question templates with an object placeholder in Appendix~\ref{sec:question-templates}. After obtaining $m$ listed objects from (1), we pick $m$ random question templates from the pool and replace the object placeholders with the listed objects $o_i$ to construct the reference questions $r_i = \{r_{i,j}\}^m_{j=1}$. We prompt the LLM to paraphrase $r_i$ to $q_i = \{q_{i,j}\}^m_{j=1}$ to increase the lexical variety of the rewritten questions $q_i$.

\subsection{Human Evaluation Guidelines}
\label{sec:human-eval-guidelines}

We conduct a human evaluation to verify and analyze the quality of the generated questions obtained from \S\ref{sec:prompt-method}, as well as measure the effectiveness of the automatic validations performed.
% Due to the distinct natures of our question generation methods, we have different human evaluation approaches for the single-turn (i.e., \textbf{generate-from-scratch}) and multi-turn prompting method (i.e., \textbf{list-then-rewrite}).
We employ three human annotators to perform the human evaluations. Detailed guidelines and examples are given prior to evaluation. We collect generated questions that are judged as valid and invalid by their automatic validation methods. Given a visual context, an image caption, a ground truth answer $\in A^\mathrm{NegP}$, and a generated question, the annotators are asked to determine whether: 1) the question is valid, 2) the question has a possible \others~answer alternative, 3) the question does not match the answer (according to both the image caption and the image), 4) the question does not match the answer (only according to the image), or 5) the question is unclear or confusing. The examples provided for each category can be seen in Appendix~\ref{sec:human-eval-category-ex}.

% \paragraph{List-then-rewrite}

% For each method, we collect the generated questions with a specification as follows: 50\% are judged as valid by the automatic evaluation, 15\% contain an object that is considered present in the image by the object matching, and the rest 35\% have a non-existent object, but is classified as a contradiction by the NLI model in the automatic evaluation.

% Given a visual context, an image caption, a ground truth answer $\in A^\mathrm{NegP}$, and a generated question, the annotators are requested to determine whether: 1) the question is valid, 2) the question contains an existent object in the image (according to both the image caption and the image), 3) the question contains an existent object in the image (only according to the image), or 4) the generated question does not make sense.
% % The examples of each category are provided in Appendix~\ref{app:}.

\begin{figure}[!t]
  \centering
  \includegraphics[width=0.825\linewidth]{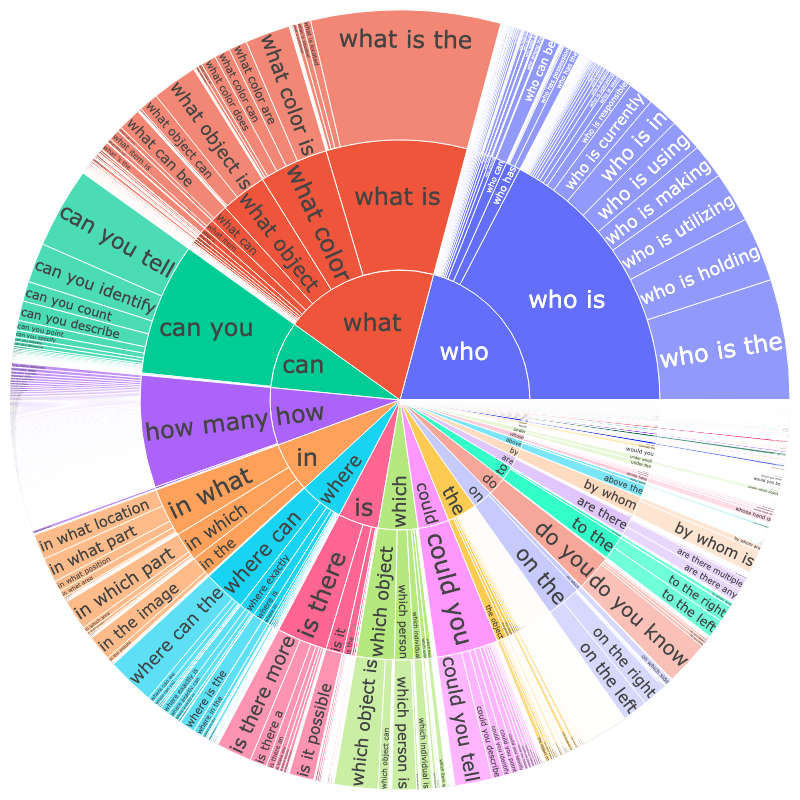}
  \caption{Distribution of NOPE's \negp~questions by their starting phrases. The arc length is proportional to the number of questions containing the word.}
  \label{fig:list-then-rewrite-sunburst}
  \vspace{-8pt}
\end{figure}

\subsection{Results and Quality Analysis}
\label{sec:data-generation-results}

Using automatic validation approaches explained in \S\ref{sec:prompt-method} and implementation details in Appendix~\ref{sec:method-impl}, we compare the capabilities of various instruction-tuned LLMs in generating \negp~VQA data. From the automatic validation results and analysis presented in Appendix~\ref{sec:auto-validation-results}, we find that employing ChatGPT yields the highest-quality generated \negp~questions by both \textbf{generate-from-scratch} and \textbf{list-then-rewrite} prompting methods, hence its use in the human evaluations. We conduct a human evaluation on randomly selected 150 generated questions from each method. For each sample, we ask 3 human experts to judge each generated question into one of the 5 options defined in \S\ref{sec:human-eval-guidelines}.
% In addition, Appendix~\ref{sec:auto-validation-results} also provides the analysis of our automatic validation's effectiveness in filtering the valid generated questions. 
% In this section, we provide the human evaluation results for each prompting method.

\begin{figure}[!t]
  \centering
  \includegraphics[width=0.95\linewidth]{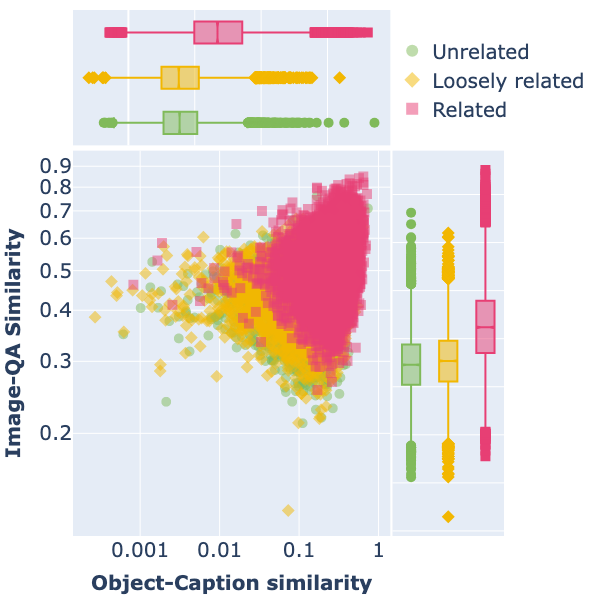}
  \caption{Object-scene relevance in the NOPE dataset. \textbf{Related} denotes ``closely related'' and \textbf{unrelated} denotes ``completely unrelated'' for brevity.}
  \label{fig:object-scene-relevance}
  \vspace{-9.25pt}
\end{figure}

\begin{table*}[!t]
    \centering
    \begin{minipage}[!b]{0.4\linewidth}
        \vspace{2.35em}
        \resizebox{\linewidth}{!}{
            \begin{tabular}{l|r|r}
                 & \multicolumn{1}{c|}{\textbf{dev}} & \multicolumn{1}{c}{\textbf{test}} \\
                 \cmidrule{2-3}
                % \multicolumn{3}{c}{} \\
                \negpbold~& \textbf{14718} & \textbf{17983} \\
                \toprule
                NOPE (\S\ref{sec:none}) & 14718 & 14773 \\
                AdVQA & 0 & 88 \\
                R-VQA & 0 & 9 \\
                TDIUC & 0 & 6 \\
                Visual7W & 0 & 1276 \\
                VQAv1 Abstract Scenes & 0 & 180 \\
                VQAv2 Balanced Real & 0 & 1651 \\
                \bottomrule
            \end{tabular}
        }
    \end{minipage}
    \hspace{2em}
    \begin{minipage}{0.4\linewidth}
        \resizebox{\linewidth}{!}{
            \begin{tabular}{l|r|r}
                 & \multicolumn{1}{c|}{\textbf{dev}} & \multicolumn{1}{c}{\textbf{test}} \\
                 \cmidrule{2-3}
                % \multicolumn{2}{l}{$\neg\mathbf{NegP}$} \\
                \textbf{\othersbold} & \textbf{14850} & \textbf{18150} \\
                \toprule
                AdVQA & 1350 & 1650 \\
                R-VQA & 2700 & 3300 \\
                TDIUC & 1350 & 1650 \\
                TextVQA & 1350 & 1650 \\
                Visual7W & 2700 & 3300 \\
                VizWiz & 1350 & 1650 \\
                VQA-Rephrasings & 1350 & 1650 \\
                VQAv1 Abstract Scenes & 1350 & 1650 \\
                VQAv2 Balanced Real & 1350 & 1650 \\
                \bottomrule
            \end{tabular}
        }
    \end{minipage}
    \caption{The data statistics of \negpbold~\textbf{(left)} and \othersbold~\textbf{(right)} subsets used in the evaluation.}
    \label{tab:nope-data-statistics}
    \vspace{-3pt}
\end{table*}

Figure~\ref{fig:simple-human-eval} shows the result of our human evaluation. For \textbf{generate-from-scratch}, only $\pm$50\% out of the subset that is judged as valid by the automatic validation is actually a valid and appropriate \negp~question according to the human annotators, and the rest is judged as incorrect by human annotators. The \textbf{list-then-rewrite} prompting approach, on the other hand, displays a significantly better question-answer generation quality with $\pm$92\% of the generated questions denoted as valid by the human annotators. This fact demonstrates that existing LLMs still fail to perform complex tasks in an end-to-end manner, while decomposing the complex tasks into several subtasks and coupling them with simple rule-based approaches can 
significantly improve the LLMs' ability to perform the complex task effectively and efficiently.

% Constructing \negp~questions require the instruction-tuned LLM to implicitly break down the task into a few intermediate steps: 1) identify a set of objects that are present according to the image caption, 2) choose an object that is not a member of that set, 3) construct a question using the non-existent object while keeping it relevant to the image caption at the same time. 12\% of the \negp~generated questions by \textbf{generate-from-scratch} include an existing object despite its presence in the respective image caption provided.

A closer look at the questions generated by the \textbf{generate-from-scratch} method shows that while LLMs usually succeed in making questions in an end-to-end manner, 12\% of the \negp~generated questions include an existing object even though this information is sufficiently provided by the image caption. Moreover, 14\% of the time, the generated questions also fail to include any objects and are overly generic, e.g., ``What is not included in this image?'', which aligns with the observations of~\cite{jang2023can, hosseini-etal-2021-understanding, ettinger2020bert, kassner2020negated} that LMs perform poorly on negation and struggles to understand that negation changes semantics. These facts show that LLMs cannot consistently perform this implicit task breakdown. From this human evaluation result, we can conjecture that the \textbf{generate-from-scratch} prompting method is not reliable and fails to elicit the LLMs' understanding of complex tasks such as question generation. Using the \textbf{list-then-rewrite} method, we generate 29.5k \negp~VQA data to build the NOPE dataset from OpenImagesV7~\cite{kuznetsova2020open}.

\subsection{Dataset Statistics}
\label{sec:none}

\paragraph{\negpbold~Question Distribution}
We cluster the generated questions into various types based on the starting n-grams in Figure~\ref{fig:list-then-rewrite-sunburst}. NOPE dataset exhibits a very broad lexical diversity of the generated questions, including variations in which the questions start with words other than the typical interrogative words (e.g., ``what'', ``where'', ``how'', etc.), such as ``Could you tell...'', ``In what location...'', ``Do you know...'', and more. This is vital to resist VL models' notorious brittleness against linguistic variations~\cite{shah2019cycle, ray2019sunny, Kervadec_2021_CVPR, whitehead2020vqap2}.

% The main question categories in our generated \negp~VQA data are as follows: 1) color attributes, e.g., ``Could you tell me the color of the jacket?''; 2) positional reasoning, e.g., ``Where exactly is the suitcase in the image?'' and ``Do you know what object is located above the counter?''; 3) counting, e.g., ``How many spoons can you find in the picture?''; and 4) object user, e.g., ``Who is seen using the computer?''.
% Additionally, our generated \negp~questions cover three different roles of the non-existent object: 1) non-existent object in question (e.g., ``Where in the image are the spices present?''), 2) non-existent reference object (e.g., ``What is the object above the grill grates?''), and 3) non-existent object in use (e.g., ``The steel water bottle is being used by whom?'').

\paragraph{Object-Scene Relevance}

Based on the descriptor used in the object listing step in \textbf{list-then-rewrite}$^{\ref{fn:diff-level}}$, the data in NOPE are divided into three categories. Figure~\ref{fig:object-scene-relevance} illustrates how these object-scene relevance descriptors of the generated \negp~VQA data correspond to the relationship between the textual semantic similarity of the selected object and the image caption, as well as the image-text semantic similarity of the image and the QA pair. We compute the textual similarity using the SentenceTransformer library\footnote{\url{https://www.sbert.net/docs/usage/semantic_textual_similarity.html}} and the image-text similarity using CLIPScore~\cite{hessel-etal-2021-clipscore}.

% Through this method, we manage to collect 30+ unique 1-gram and 90+ unique 2-gram starting phrases (with a number of occurrences at least 10 times), a lot more diverse than previous works (details in Appendix~\ref{sec:existing-question-diversity}, to support a greater variety of linguistic choices. We provide the answer distribution of the most common 2-gram starting phrases in Figure~\ref{fig:list-then-rewrite-ans-dist}.

% \begin{figure}[t]
%   \centering
%   \includegraphics[width=1.0\linewidth]{images/vqa-h-list-then-rewrite-ans-dist.png}
%   \caption{Answer distribution of \negp~questions with the most common starting 2-grams.}
%   \label{fig:list-then-rewrite-ans-dist}
% \end{figure}

% \paragraph{Semantic Similarity of Existent and Non-Existent Objects}

% \paragraph{Correlation Between NLI and Human Judgement}

% \holy{Statistics, whether image and QA are related to each other, comparison with existing NP questions, etc.}

%%%%%%%%%%%%%%%%%%%%

\section{Experimental Settings}
\label{sec:benchmark}

The object hallucination benchmark consists of the validation and test sets of 10 VQA corpora, including NOPE (\S\ref{sec:none}) with balanced object-scene relevance proportions. It displays the comparison between incorrectness and object hallucination over various baselines, which serves as a foundation for assessing object hallucination in addition to the standard incorrectness in 10 VL models.

% \begin{table}[t]
%     \centering
%     \begin{tabular}{l|r|r}
%         % \multicolumn{2}{l}{\textbf{Test}} \\
%         % \toprule
%         % \multicolumn{2}{l}{$\mathbf{NegP}$} \\
%          & \multicolumn{1}{c|}{\textbf{dev}} & \multicolumn{1}{c}{\textbf{test}} \\
%          \cmidrule{2-3}
%         % \multicolumn{3}{c}{} \\
%         \negpbold~& \textbf{14718} & \textbf{17983} \\
%         \toprule
%         NONE (\S\ref{sec:none}) & 14718 & 14773 \\
%         AdVQA & 0 & 88 \\
%         R-VQA & 0 & 9 \\
%         TDIUC & 0 & 6 \\
%         Visual7W & 0 & 1276 \\
%         VQAv1 Abstract Scenes & 0 & 180 \\
%         VQAv2 Balanced Real & 0 & 1651 \\
%         % \midrule
%         \multicolumn{3}{c}{} \\
%         % \multicolumn{2}{l}{$\neg\mathbf{NegP}$} \\
%         \textbf{$\neg\mathbf{NegP}$} & \textbf{14850} & \textbf{18150} \\
%         \toprule
%         AdVQA & 1350 & 1650 \\
%         R-VQA & 2700 & 3300 \\
%         TDIUC & 1350 & 1650 \\
%         TextVQA & 1350 & 1650 \\
%         Visual7W & 2700 & 3300 \\
%         VizWiz & 1350 & 1650 \\
%         VQA-Rephrasings & 1350 & 1650 \\
%         VQAv1 Abstract Scenes & 1350 & 1650 \\
%         VQAv2 Balanced Real & 1350 & 1650 \\
%         \bottomrule
%     \end{tabular}
%     \caption{NOPE benchmark data statistics.}
%     \label{tab:nope-data-statistics}
% \end{table}

\subsection{Datasets}
\label{sec:nope-datasets}

% NOPE (\textbf{N}egative \textbf{O}bject \textbf{P}resence \textbf{E}valuation) consists of the validation and test sets of 10 different VQA corpora, including the NONE dataset in \S\ref{sec:none} with balanced object-scene relevance proportions. 

\begin{figure}[!t]
  \centering
  \includegraphics[width=0.95\linewidth]{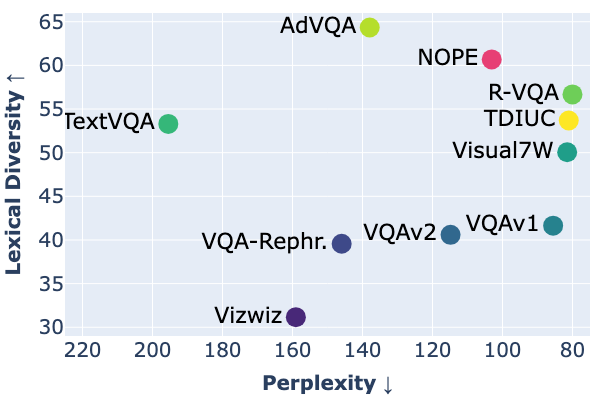}
  \vspace{-8pt}
  \caption{Question quality in the benchmark in terms of lexical diversity and fluency.}
  \label{fig:question-quality}
  \vspace{-8pt}
\end{figure}

Table~\ref{tab:nope-data-statistics} describes the data distribution of the dev and test sets of the benchmark. Each set respectively comprises $\sim$30k and $\sim$36k data, maintaining near-balanced proportions of \negp~and \others~data. To ensure the quality of the visual questions in the benchmark, we also analyze the lexical diversity and the fluency of the comprising datasets, which are useful to assert a robust evaluation using questions that are linguistically diverse and coherent. In Figure~\ref{fig:question-quality}, we show that the datasets whose data construction utilizes automatic question generation, i.e., NOPE and TDIUC, have comparable lexical diversity and fluency to the other datasets, which entirely rely on question generation by human annotators.

For lexical diversity, we employ length-agnostic lexical diversity metrics, i.e., moving average type-token ratio (MATTR)~\cite{covington2010mattr}, measure of textual lexical diversity (MTLD)~\cite{mccarthy2005mtld}, and hypergeometric distribution diversity (HDD)~\cite{mccarthy2007vocd,mccarthy2010mtldhdd}, and average them. We use LexicalRichness~\cite{shen2021accuracybias,shen2022lexicalrichness} v0.5.0\footnote{\url{https://pypi.org/project/lexicalrichness/}} to calculate these metrics. We also employ a large pre-trained LM GPT-Neo~\cite{black2021gpt-neo} with 2.7B parameters
% \footnote{\url{https://huggingface.co/EleutherAI/gpt-neo-2.7B}}
to compute the perplexity of the questions, which is often used as a measure of both lexical diversity~\cite{lewis2017deal,tevet2021evaluating} and fluency~\cite{fan2018hierarchical,wang2019paperrobot,cahyawijaya2021indonlg,anonymous2023nusawrites}.

\subsection{Baselines}
\label{sec:nope-baselines}

For the baselines in our benchmark, we employ various vision-language model architectures on the benchmark in both zero-shot \& few-shot and fine-tuned fashion. For the fine-tuned setting, we utilize five models: 1) OFA~\cite{wang2022ofa}, which unifies architectures, tasks, and modalities by formulating a unified sequence-to-sequence abstraction via handcrafted instructions to achieve task agnosticism; 2) and 3) BLIP~\cite{li2022blip}, which incorporates two key contributions, i.e., multimodal mixture of encoder-decoder (MED) to operate as either a unimodal encoder, an image-grounded text encoder, or an image-grounded text decoder, and CapFilt as a new dataset bootstrapping method for learning from noisy image-text pairs; 4) ALBEF~\cite{li2021align}, which is trained using momentum distillation to improve learning from noisy web data; 5) GIT~\cite{wang2022git}, which employs an image encoder and a text decoder pre-trained using a language modeling objective to map the input image to its corresponding description.

For the zero-shot setting, we employ: 1) BLIP-2~\cite{li2023blip}, which utilizes a scalable multimodal pre-training method to enable any LLMs to ingest and understand images; 2) and 3) PromptCap~\cite{hu2022promptcap}, which is trained to generate captions that help downstream LMs answer visual questions; 4) InstructBLIP~\cite{instructblip}, which is an instruction-tuned version of BLIP-2 on various tasks including VQA. We also employ 5) OpenFlamingo~\cite{alayrac2022Flamingo, anas_awadalla_2023_7733589}, which is an open-source version of a large pre-trained VL model specialized in few-shot prompting, in the two-shot setting. Table~\ref{tab:baselines} provides the model and data sizes of the baselines and Appendix~\ref{sec:baseline-details} lists the model variants.

\begin{table}[t]
    \centering
    \resizebox{0.825\linewidth}{!}{
        \begin{tabular}{lrr}
            \toprule
            \multirow{2}{*}{\textbf{}} & \multirow{2}{*}{\makecell{\textbf{Model} \\ \textbf{size}}} & \multirow{2}{*}{\makecell{\textbf{\# Pre-train} \\ \textbf{images}}} \\
            % \cmidrule{4-4} \cmidrule{5-6}
             & & \\
            \toprule
            \multicolumn{3}{>{\columncolor[gray]{.9}}c}{\textbf{\textit{Zero-shot \& Few-shot}}} \\
            PromptCap$_{BASE}$ & 696M & 34M \\
            PromptCap & 3B & 34M \\
            BLIP-2 & 3.8B & 129M \\
            OpenFlamingo & 9B &  $\sim$2.5B \\
            \midrule
            \multicolumn{3}{>{\columncolor[gray]{.9}}c}{\textbf{\textit{VQA fine-tuned}}} \\
            OFA & 929M & 34M \\
            BLIP & 385M & 129M \\
            BLIP$_{CapFilt-L}$ & 385M & 129M \\
            ALBEF & 628M & 14M \\
            GIT$_{LARGE}$ & 347M & 1.4B \\
            InstructBLIP$_{FLAN_{XL}}$ & 3.8B & 129M+ \\
            \bottomrule
        \end{tabular}
    }
    \caption{VL baseline models in the benchmark.}
    \label{tab:baselines}
    \vspace{-5pt}
\end{table}

\begin{table*}[t]
    \centering
    \resizebox{0.98\linewidth}{!}{
    \begin{tabular}{l cc ccccccccc}
        \toprule
        \multirow{3}{*}{\textbf{}} & \multicolumn{2}{c}{\textbf{\othersbold~test (\%)}} & \multicolumn{7}{c}{\textbf{\negpbold~test (\%)}} \\
        \cmidrule(lr){2-3} \cmidrule(lr){4-10}
        & \multicolumn{2}{c}{\textbf{Overall}} & \multicolumn{2}{c}{Existing datasets} & \multicolumn{2}{c}{NOPE test (\S\ref{sec:none})} & \multicolumn{3}{c}{\textbf{Overall}} \\
        \cmidrule(lr){2-3} \cmidrule(lr){4-5} \cmidrule(lr){6-7} \cmidrule(lr){8-10}
        & \multirow{1}{*}{Acc.} & \multirow{1}{*}{METEOR} & \multirow{1}{*}{Acc.} & \multirow{1}{*}{METEOR} & \multirow{1}{*}{Acc.} & \multirow{1}{*}{METEOR} & \multirow{1}{*}{$\mathrm{NegP}$ Acc.} & \multirow{1}{*}{Acc.} & \multirow{1}{*}{METEOR} \\
        \toprule
        \multicolumn{10}{>{\columncolor[gray]{.9}}c}{\textbf{\textit{Zero-shot \& few-shot}}} \\
        PromptCap$_{BASE}$ & 30.18 & 21.45 & 2.87 & 3.05 & 0.21 & 0.29 & 0.95 & 0.68 & 0.78 \\
        PromptCap & \ul{32.69} & 22.66 & 3.61 & 2.20 & 0.42 & 0.56 & 1.67 & 0.99 & 0.85 \\
        BLIP-2 & 19.84 & 17.94 & \ul{4.39} & 1.49 & \ul{2.11} & \ul{1.22} & \ul{5.25} & \ul{2.51} & 1.27 \\
        OpenFlamingo & 14.29 & \textbf{24.32} & 0.09 & \ul{7.96} & 0.00 & 0.08 & 0.02 & 0.02 & \ul{1.49} \\
        \midrule
        \multicolumn{10}{>{\columncolor[gray]{.9}}c}{\textbf{\textit{VQA fine-tuned}}} \\
        OFA & 29.43 & 17.06 & 3.24 & 4.10 & 2.75 & \textbf{9.11} & 8.21 & 2.84 & 8.21 \\
        BLIP & 23.27 & 12.07 & 5.95 & 5.12 & 1.60 & 3.63 & 6.48 & 2.38 & 3.90 \\
        BLIP$_{CapFilt-L}$ & 23.28 & 12.08 & 5.95 & 5.12 & 1.60 & 3.61 & 6.47 & 2.37 & 3.88 \\
        ALBEF & 16.33 & 21.87 & 19.31 & \textbf{26.31} & 1.86 & 6.76 & 8.18 & 4.98 & \textbf{10.25} \\
        GIT$_{LARGE}$ & \textbf{41.00} & 21.75 & \textbf{34.89} & 20.43 & 4.00 & 5.90 & \textbf{17.92} & \textbf{9.51} & 8.49 \\
        InstructBLIP & 40.62 & \ul{22.55} & 21.40 & 13.50 & \textbf{5.08} & 5.19 & 17.69 & 7.99 & 6.67 \\
        \bottomrule
    \end{tabular}}
    \caption{Weighted model performances on the test set of the benchmark. Errors made on \textcolor{non_np_color}{\others} VQA data represent incorrectness, while errors made on \textcolor{np_color}{\negp} VQA data represent object hallucination. \textbf{Bold} and \ul{underline} denote the best performances overall and in the group, respectively.}
    \label{tab:nope-benchmark-test-results}
    % \vspace{-4pt}
\end{table*}

\subsection{Evaluation Settings}
\label{sec:evaluation-settings}

For both \negp~and \others, we compute accuracy and METEOR~\cite{banerjee2005meteor} to measure the performance of vision-language models on the benchmark. While accuracy measures the performance based on an exact match between the generated answer and the ground truth answer, METEOR caters to partial (i.e., unigram) matches by computing a score for this matching using a combination of unigram-precision, unigram-recall, and alignment between the unigrams in the generated answer and ground truth answer. Additionally, for \negp, we employ a rule-based accuracy, referred to as \negp~accuracy, which focuses on determining whether the generated answer is a negative indefinite pronoun (i.e., $\in A^\mathrm{NegP} = \{"none", "nothing", "nowhere", "zero", "0", \\ "no\;one", "nobody", "neither"\}$) or not. All scores are computed per task and then the weighted averages according to each task size are retrieved.

\section{Results}
\label{sec:nope-results}

\begin{figure}[t]
    \centering
    \includegraphics[width=1.0\linewidth, trim={0 1cm 0 0.5cm}]{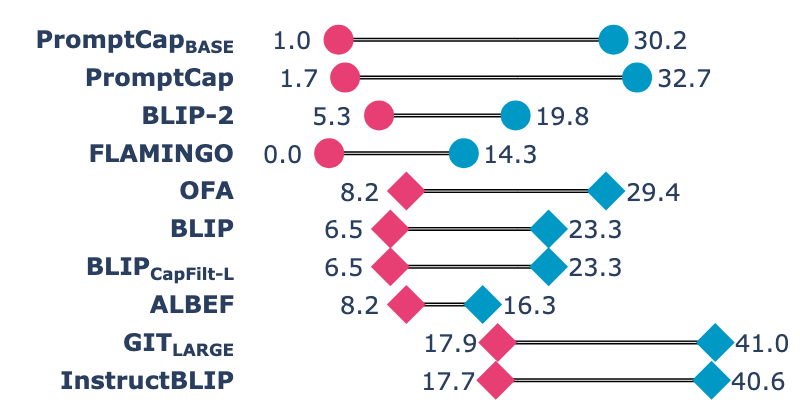}
    \caption{All baselines consistently score lower on \textcolor{np_color}{\negp} (\%$\mathrm{NegP}$ Acc.) than \textcolor{non_np_color}{\others} (\%Acc.).}
    \label{fig:nope-benchmark-test-results}
    % \vspace{-5pt}
\end{figure}

We present the results on the test set of the benchmark in Table~\ref{tab:nope-benchmark-test-results}. Examples of object hallucination are in Appendix~\ref{sec:obj-hall-examples}.
% Figure~\ref{fig:nope-benchmark-test-results}.
While the VQA-finetuned baselines are slightly better at \negp~and comparable to the zero-shot \& few-shot baselines on \others, as in Figure~\ref{fig:nope-benchmark-test-results}, we observe that all zero-shot and VQA-finetuned baselines notably perform much worse on \negp~tasks that \others~with the averaged discrepancies of $\pm$22\% and $\pm$18\% accuracy, respectively. This demonstrates that all baselines are more vulnerable and susceptible to object hallucination than standard incorrectness. In addition, less incorrectness does not entail less object hallucination. For instance, PromptCap$_{BASE}$, PromptCap, and BLIP have lower scores on \negp~than ALBEF despite outperforming it on \others~setting. It also means that existing evaluations that solely utilize \others~cases cannot effectively capture the models' risk of object hallucination.

Another point that we observe is, GIT outperforms the other baselines on both \negp~and \others~data, as well as manages to surpass much bigger models (e.g., InstructBLIP and Flamingo), showing that GIT is more robust against both object hallucination and general incorrectness, despite being the smallest in size (Table \ref{tab:baselines}) and having a simple architecture. This achievement could be attributed to its substantial number of pre-training images, which is an order of magnitude larger than those of the other baselines. This also aligns with~\cite{hoffmann2022training}, in which for the same compute budget, a smaller model trained on more data outperforms a larger model trained on fewer data and achieves more optimal performance.

\section{Analysis and Discussions}

\subsection{Object hallucination and lexical diversity}

Table~\ref{tab:nope-benchmark-test-results} also show that \negp~performance scores on existing datasets are significantly higher than on NOPE across the metrics, indicating that object hallucination is more likely to occur when the models attempt to solve the questions in NOPE. This is mainly due to the NOPE dataset having a relatively higher lexical diversity compared to the other \negp~corpora, which are mostly composed of VQAv2 and Visual7W (see in Figure~\ref{fig:question-quality}). This also aligns with the fact that \negp~model performances have a strong negative Pearson correlation with the lexical diversity measures ($r=\{-0.8, -0.66, -0.65, -0.7\}$ for METEOR and HDD, MTLD, MATTR, perplexity) and proves that corpora with higher lexical diversity (e.g., NOPE) provide more challenging \negp~VQA problems to assess object hallucination.

\begin{figure}[t]
    \centering
    \includegraphics[width=1.0\linewidth, trim={0 0.5cm 0 0}]{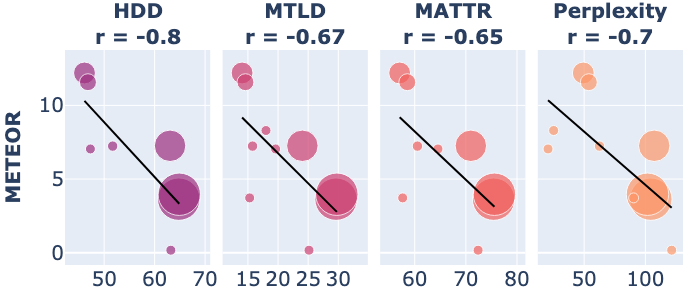}
    \caption{VL models are more prone to object hallucination on lexically diverse \textcolor{np_color}{\negp} VQA data. Dot size represents dataset size (\S\ref{sec:nope-datasets}).}
    \label{fig:corr-obj-hallucination-ld}
    % \vspace{-5pt}
\end{figure}

\subsection{Object hallucination and language bias}

\begin{figure}[t]
    \centering
    \includegraphics[width=0.92\linewidth, trim={1.8cm 1.8cm 0 0cm}]{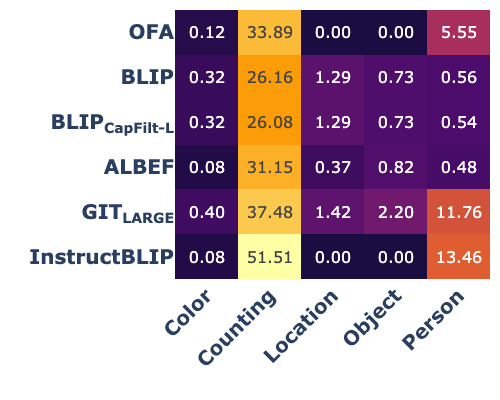}
    \caption{\textcolor{np_color}{\negp} performance of VQA fine-tuned baselines over different question types.}
    \label{fig:nope-acc-per-q-type}
    \vspace{-6pt}
\end{figure}

As shown in Figure~\ref{fig:nope-acc-per-q-type}, among 5 \negp~question types, all VQA-finetuned VL models fail on \negp~questions about color (e.g., ``What is the color of...?''), object (e.g., ``What is the object beside...?''), and location (e.g., ``Where is...?''), while most VL models tend to hallucinate less on \negp~questions about counting (e.g., ``How many...?'') and person (e.g., ``Who is using...?''). A similar trend is observed for the zero-shot \& few-shot baselines. We further inspect these two categories and find out that their answer scopes are of a smaller scope than the others in the training data. For instance, the answers to counting questions are often numbers $\leq$ 5, and the answers to the person questions are often the generic "man", "woman", "person", "people", and others which have fewer variations compared to object types, color names, or absolute and relative places. These facts suggest that existing VL models have a strong language bias~\cite{kv2020langbias,niu2021langbias,wu2022langprior} toward certain question types, which result in acceptable \negp~performances on those question types. Nevertheless, language bias does not solve object hallucination and even might make it worse, due to the VL models having weak visual grounding skills to verify the answer to the visual context, which might lead to errors on both \negp~and \others~questions.

\subsection{Object hallucination and object-scene relevance}

\begin{figure}[t]
    \centering
    \includegraphics[width=0.825\linewidth, trim={0 2cm 0 0.5cm}]{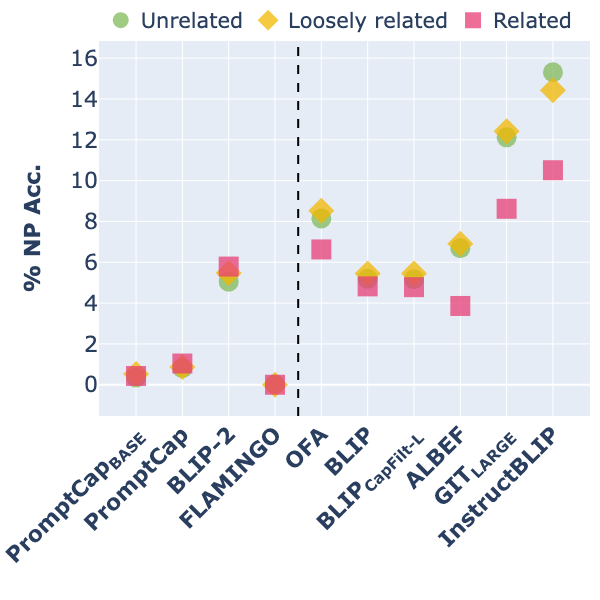}
    \caption{\textcolor{np_color}{\negp} performance of \textbf{(left}) zero-shot \& few-shot and \textbf{(right)} VQA fine-tuned baselines per object-scene relevance.}
    \label{fig:nope-acc-per-diff}
    \vspace{-5pt}
\end{figure}

As shown in Figure~\ref{fig:nope-acc-per-diff}, all VQA fine-tuned models perform lower when the object is closely related to the scene compared to when the object is loosely related or unrelated. This indicates that VL models have some degree of understanding \negp~based on the relevance of the object in question with the scene. Although this helps VL models to understand about objects better in some cases, this also causes VL models to hallucinate more on objects that are relevant to the scene~\cite{rohrbach2018object,kayhan2021hallu-od,dai2022plausible}. 
% Solving this problem will likely correlate to the difficulty in ObjectNet~\cite{mayo2023how}. 
On the other hand, the performance on loosely related or unrelated objects tend to be similar, which aligns with the similarity analysis provided in Figure~\ref{fig:object-scene-relevance}. In contrast, for zero-shot \& few-shot baselines, the differences between object-scene relevance are less apparent. However, in general, the \negp~scores are also very low, except for BLIP-2, which suggests that most zero-shot models do not have an adequate understanding of \negp.

% \subsection{Dummy}
% \dummy{\lipsum[1]}

% \subsection{Does learning more \negpbold~help?}

% Provided the scalable method to generate synthetic data, we further analyze the effect of learning more \negp~data to VL models. Specifically, we experiment with fine-tuning BLIP with the validation set of NOPE in three settings by using: 1) only $\neg \mathrm{NegP}$ data, 2) only $\neg \mathrm{NegP}$ data,, and 3)  \negp~and evaluate on the test set,  As shown in Figure~\ref{fig:learning-np}, BLIP without further fine-tuning is outperformed by the models trained

\section{Conclusion}

We have addressed the critical issue of object hallucination in VL models, which has been lacking a general measurement. We have introduced NOPE to assess object hallucination in VL models, investigating the discernment of objects' non-existence in visual questions by 10 state-of-the-art VL models, alongside their standard performances. Additionally, we have presented a cost-effective and scalable method for generating high-quality synthetic data with over 90\% validity to overcome the severe underrepresentation of \negp~cases. Through our comprehensive experiments, we have demonstrated that no VL model is exempt from object hallucination, highlighting their lack of understanding of negative object presence. Furthermore, we have identified lexical diversity, question type, and the relevance of the object to the visual scene as influential factors impacting VL models' susceptibility to object hallucination. These findings provide valuable insights into the assessment of object hallucination in VL, thereby paving the way for the future development of enhanced VL models.

% In this work, we highlight the critical issue of object hallucination in VL models and its lack of a formal definition and quantitative evaluation. We introduce Negative Object Presence Evaluation (NOPE) to assess object hallucination and standard incorrectness in VL models, which investigates the ability of 10 state-of-the-art models to discern objects' non-existence in visual questions in addition to the standard performances. We also present a cost-effective and scalable method for generating high-quality synthetic data with >90\% validity to address the extreme underrepresentation of \negp~cases. Our findings demonstrate that no VL model is immune to object hallucination, and we show that lexical diversity, question type, and the relevance of the object in question to the visual scene have a considerable impact to VL models' susceptibility to object hallucination. Our research offers valuable insights into the assessment and implications of object hallucination in VL, paving the way for improved VL models in the future.

\section{Limitation and Future Work}

\paragraph{Evaluation Metrics for Object Hallucination}
In this work, we show three metrics to measure object hallucination and incorrectness, i.e., the exact match accuracy, METEOR, and \negp~accuracy. Nevertheless, in some cases, these metrics fail to capture some equivalent answer that has the same semantic meaning. For example, given an \negp~question ``Where is the spoon in the picture?'' with the corresponding label ``Nowhere'', a system that answers with ``There is no spoon in the picture'' will get 0 scores on these three metrics, despite the answer is actually correct. We argue that the limitation of the existing metrics might hinder further research in alleviating object hallucination and we expect future works to focus on developing better metrics for measuring object hallucination.

\paragraph{Object Hallucination Outside of \negpbold}

Since object hallucination refers to an effect (i.e., generating non-existent objects) and not a cause, our measurement of object hallucination is limited to \negp~cases, in which a VL model unfaithfully infers a supposedly non-existent object as existent in the visual context. For cases where a VL model provides an incorrect answer to \others~VQA, the fine line between misclassification and object hallucination has not yet been defined.

\paragraph{Performances on Full \othersbold~Test Sets}
In order to observe the incorrectness of VL models on \others~on various datasets, we compose a balanced set of $\sim$15k data in our dev split and $\sim$18k data in our test split from diverse VQA corpora. Obtaining the full performance on each of the source datasets requires re-running the baselines on the full test sets of each source dataset.

\section{Ethics Statement}

This research on object hallucination in vision-language models aims to improve the reliability and faithfulness of these models, which have significant applications in various fields such as healthcare and autonomous driving. We acknowledge the potential impact of our findings and commit to promoting responsible and ethical use of these models. We recognize that such models have the potential to perpetuate biases and stereotypes, and we have taken steps to mitigate this risk. For instance, we ensured that the synthetic data used in this study was generated in a manner that respects privacy and does not perpetuate biases or stereotypes. Furthermore, we recognize the importance of transparency and accountability in the development and use of these models. Therefore, we commit to sharing our findings and methodologies openly and making them accessible to the wider research community. We also acknowledge that these models can have unintended consequences and commit to ongoing monitoring and evaluation of their impact. Finally, we recognize that the development and use of these models must be guided by ethical principles that prioritize human well-being and social responsibility. We are committed to upholding these principles and contributing to the development of responsible and ethical practices in the field of vision-language modeling.

\section*{Acknowledgements}
This work has been partially funded by PhD Fellowship Award, the Hong Kong University of Science and Technology; PF20-43679 Hong Kong PhD Fellowship Scheme, Research Grant Council, Hong Kong; and the National Research Foundation, Singapore under its AI Singapore Programme.

% This document has been adapted by Yue Zhang, Ryan Cotterell and Lea Frermann from the style files used for earlier ACL and NAACL proceedings, including those for 
% ACL 2020 by Steven Bethard, Ryan Cotterell and Rui Yan,
% ACL 2019 by Douwe Kiela and Ivan Vuli\'{c},
% NAACL 2019 by Stephanie Lukin and Alla Roskovskaya, 
% ACL 2018 by Shay Cohen, Kevin Gimpel, and Wei Lu, 
% NAACL 2018 by Margaret Mitchell and Stephanie Lukin,
% Bib\TeX{} suggestions for (NA)ACL 2017/2018 from Jason Eisner,
% ACL 2017 by Dan Gildea and Min-Yen Kan, NAACL 2017 by Margaret Mitchell, 
% ACL 2012 by Maggie Li and Michael White, 
% ACL 2010 by Jing-Shin Chang and Philipp Koehn, 
% ACL 2008 by Johanna D. Moore, Simone Teufel, James Allan, and Sadaoki Furui, 
% ACL 2005 by Hwee Tou Ng and Kemal Oflazer, 
% ACL 2002 by Eugene Charniak and Dekang Lin, 
% and earlier ACL and EACL formats written by several people, including
% John Chen, Henry S. Thompson and Donald Walker.
% Additional elements were taken from the formatting instructions of the \emph{International Joint Conference on Artificial Intelligence} and the \emph{Conference on Computer Vision and Pattern Recognition}.

\bibliography{emnlp2023}
\bibliographystyle{acl_natbib}

\clearpage
\appendix
\onecolumn

\section{Prompt Templates}
\label{sec:prompt-templates}

We provide the prompt templates and examples for the \textbf{generate-from-scratch} and \textbf{list-then-rewrite} methods in Table~\ref{tab:generate-from-scratch-prompt-template} and Table~\ref{tab:list-then-rewrite-prompt-template}, respectively.

\begin{table*}[h]
    \centering
    \resizebox{\linewidth}{!}{
    \begin{tabular}{p{16cm}}
        \toprule
        \textbf{Template 1} \\
        Create a question beginning with "<INTERROGATIVE\_WORD>" from this image caption: "<IMAGE\_CAPTION>" with an answer of "<ANSWER>". \\
        \textbf{Example:} Create a question beginning with "who" from this image caption: "This image consists of an airplane in the air. On which, we can see the text. In the background, there is sky." with an answer of "nobody". \\
        \textbf{Generated question:} Who is in the airplane in this image? \\
        \midrule
        \textbf{Template 2} \\
        Turn the text "<IMAGE\_CAPTION>" into a "<INTERROGATIVE\_WORD>" question with an answer "<ANSWER>": \\
        \textbf{Example:} Turn the text "This is an outside view. Here I can see the cars on the ground and there are two persons standing. In the background there is a tent, shelter and trees. At the top of the image there is the sky. At the bottom there is grass." into a "how many" question with an answer "none": \\
        \textbf{Generated question:} How many buildings are visible in the outside view? \\
        \midrule
        \textbf{Template 3} \\
        Given an answer "<ANSWER>", generate an appropriate question starting with "<INTERROGATIVE\_WORD>" from the text "<IMAGE\_CAPTION>". \\
        \textbf{Example:} Given an answer "nowhere", generate an appropriate question starting with "where" from the text "In this picture, we see a woman. She is staring at something and she might be posing for the photo. We see a nail is pierced on her face. She might be taking bath. In the background, we see a wall and a white color object which looks like a bathtub.". \\
        \textbf{Generated question:} Where is the ideal location to put a nail on a person's face while they are taking a bath? \\
        \midrule
        \textbf{Template 4} \\
        Write a question that begins with "<INTERROGATIVE\_WORD>" and is answered by "<ANSWER>" based on the text "<IMAGE\_CAPTION>". \\
        \textbf{Example:} Write a question that begins with "what" and is answered by "nothing" based on the text "In this image I can see food item in a cooking vessel, which is on the stove. In the background there are some objects.". \\
        \textbf{Generated question:} What can you see in the background of the image? \\
        \midrule
        \textbf{Template 5} \\
        Rewrite the statement "<IMAGE\_CAPTION>" as a question that starts with interrogative phrase "<INTERROGATIVE\_WORD>" and has an answer of "<ANSWER>". \\
        \textbf{Example:} Rewrite the statement "In the image I can see food item in the bowl. The background of the image is blurred." as a question that starts with interrogative phrase "which" and has an answer of "neither". \\
        \textbf{Generated question:} Which part of the image is not blurred and which object is visible in the bowl? \\
        \bottomrule
    \end{tabular}}
    \caption{Prompt templates used in the \textbf{generate-from-scratch} (\S\ref{sec:prompt-method}) and their corresponding examples.}
    \label{tab:generate-from-scratch-prompt-template}
\end{table*}

\begin{table*}[h]
    \centering
    \begin{tabular}{p{16cm}}
        \toprule
        \textbf{Object Listing Template} \\
        Description: \\
        "<IMAGE\_CAPTION>" \\
        \\
        List 10 tangible objects that are closely related to the description, but are not mentioned there. Also avoid these objects: <IMAGE\_LABELS>. \\
        \textbf{Example:} \\
        Description: \\
        "In this picture we can see some food products in the glass jars." \\
        \\
        List 10 tangible objects that are closely related to the description, but are not mentioned there. Also avoid these objects: Vegetable. \\
        \textbf{Response:} \\
        1. Spices, 2. Canned fruits, 3. Nuts, 4. Sugar, 5. Coffee beans, 6. Honey, 7. Pasta, 8. Chocolates, 9. Tea leaves, 10. Jams and preserves \\
        \midrule
        \textbf{Question Rewriting Template} \\
        Question templates: \\
        <REFERENCE\_QUESTIONS\_BASED\_ON\_FIRST\_TURN\_ANSWER> \\
        \\
        Paraphrase each of these <NUMBER> questions while keeping the sentence as a question with a question mark. \\
        \textbf{Example}: \\
        Question templates: \\
        1. Who is using the object above the spices? \\
        2. What object is located under the canned fruits? \\
        3. Who is holding the object on the right of the nuts? \\
        4. What is the color of the sugar? \\
        5. Who is using the coffee beans? \\
        6. What object is located on the left of the honey? \\
        7. What object is located on the left of the pasta? \\
        8. What object is located on the left of the chocolates? \\
        9. What is the color of the tea leaves? \\
        10. Where is the jams and preserves located in the image? \\
        \\
        Paraphrase each of these 10 questions while keeping the sentence as a question with a question mark. \\
        \textbf{Generated question:} \\
        1. Who is the user of the spices in the picture? \\
        2. Which product can be found below the canned fruits? \\
        3. Who is the individual holding the object to the right of the nuts? \\
        4. Can you tell me the color of the sugar in the glass jars? \\
        5. Who might be utilizing the coffee beans shown in the image? \\
        6. What is the item on the right side of the honey? \\
        7. What comes up on the left of the pasta in the photograph? \\
        8. What is situated on the left of the chocolates in the picture? \\
        9. What is the color of the tea leaves displayed in the glass jars? \\
        10. Where exactly can the jams and preserves be found in this image? \\
        \bottomrule
    \end{tabular}
    \caption{Prompt templates used in the \textbf{list-then-rewrite} (\S\ref{sec:prompt-method}) and their corresponding examples.}
    \label{tab:list-then-rewrite-prompt-template}
\end{table*}

\clearpage

\section{Reference Question Templates}
\label{sec:question-templates}

Table~\ref{tab:question-template} presents the pool of question templates used to automatically build the reference questions for the \textbf{list-then-rewrite} in \S\ref{sec:prompt-method}.

\begin{table*}[ht]
    \centering
    \resizebox{1.0\linewidth}{!}{
    \begin{tabular}{l p{12cm} p{2.5cm}}
        \toprule
        \textbf{No} & \textbf{Question template} & \textbf{$\mathbf{NegP}$ answer} \\
        \midrule
        1 & What is the color of the <OBJECT>? & none / nothing \\
        2 & What object is located above / under / on the left of / on the right of the <OBJECT>? & none / nothing \\
        3 & Where is the <OBJECT> located in the image? & nowhere \\
        4 & How many <OBJECT> are there in the image? & zero / 0 / none \\
        5 & Who is holding / using the <OBJECT>? & no one / nobody \\
        6 & Who is holding / using the object above / under / on the left of / on the right of the <OBJECT>? & no one / nobody \\
        \midrule
    \end{tabular}}
    \caption{Question templates utilized to construct the reference questions for the question rewriting step in the \textbf{list-then-rewrite} prompting methodology in \S\ref{sec:prompt-method}.}
    \label{tab:question-template}
\end{table*}

\section{Automatic Validation Methodologies of \negpbold~VQA Data Generation}
\label{sec:auto-validation-methods}

\paragraph{Generate-from-scratch}

To ensure the validity of $q_i$, we use a model fine-tuned on natural language inference (NLI) to determine whether a generated question $q_i$ and answer $a_i$ pair (i.e., hypothesis) logically entails its corresponding image caption $c_i$ (i.e., premise). We also utilize a fine-tuned binary classifier to determine whether a generated question $q_i$ and answer $a_i$ pair fits a given visual context $v_i$. If the question $q_i$ and answer $a_i$ pair is true (entailment) or undetermined (neutral) given $c_i$ as well as matches with $v_i$, then the generated question $q_i$ is judged as valid by the automatic validation.

\paragraph{List-then-rewrite}

For the automatic validation of a listed object $o_{i,j}$, we extract lemmatized noun tokens from its corresponding image caption $c_i$ and obtain the object names from $l_i$ as the objects present in $v_i$. If $o_{i,j}$ does not match with any of the extracted objects, then $o_{i,j}$ is a valid non-existent object. For the automatic validation of a generated question $q_{i,j}$, if $q_{i,j}$ does not contradict its respective reference question $r_{i,j}$, then the generated question $q_{i,j}$ is considered valid.

\section{Implementation Details of \negpbold~VQA Data Generation}
\label{sec:method-impl}

We implement \S\ref{sec:prompt-method} with the following LLMs that employ: 1)  multi-task prompted fine-tuning, i.e., \textbf{BLOOMZ}~\cite{muennighoff2022crosslingual} and \textbf{T0}~\cite{sanhmultitask}; 2) instruction meta-learning, i.e., \textbf{OPT-IML}~\cite{iyer2022opt}; 3) synthetic self-instruct, i.e., \textbf{Alpaca}~\cite{wang2022self}; 4) instruction~\cite{wei2022finetuned} and chain-of-thought fine-tuning~\cite{weichain}, i.e., \textbf{FLAN T5} and \textbf{FLAN Alpaca}~\cite{chung2022scaling}; 5) multi-task instruction pre-training, i.e., \textbf{ChatGLM}~\cite{zeng2023glm-130b}; 6) conversation-style instruction tuning and reinforcement learning with human feedback (RLHF)~\cite{christiano2017deep, stiennon2020learning}, i.e., \textbf{ChatGPT (GPT-3.5)}. More details are presented in Table~\ref{tab:llm-details}.

\begin{table*}[ht]
    \centering
    \resizebox{1.0\linewidth}{!}{
    \begin{tabular}{l l l l l} %p{4cm} p{8cm}}
        \toprule
        \textbf{No} & \textbf{Model} & \textbf{Size} & \textbf{References} & \textbf{Access} \\
        \toprule
        1 & BLOOMZ (3B) & 3B & \cite{muennighoff2022crosslingual, scao2022bloom} & \url{https://huggingface.co/bigscience/bloomz-3b} \\
        2 & BLOOMZ (7.1B) & 7.1B & \cite{muennighoff2022crosslingual, scao2022bloom} & \url{https://huggingface.co/bigscience/bloomz-7b1} \\
        3 & T0 & 3B & \cite{sanhmultitask} & \url{https://huggingface.co/bigscience/T0_3B} \\
        4 & OPT-IML & 1.3B & \cite{iyer2022opt, zhang2022opt} & \url{https://huggingface.co/facebook/opt-iml-max-1.3b} \\
        5 & Alpaca & 7B & \cite{wang2022self, touvron2023llama} & \url{https://huggingface.co/chavinlo/alpaca-native} \\
        6 & FLAN T5 XL & 3B & \cite{chung2022scaling, 2020t5} & \url{https://huggingface.co/google/flan-t5-xl} \\
        7 & FLAN T5 XXL & 11B & \cite{chung2022scaling, 2020t5} & \url{https://huggingface.co/google/flan-t5-xxl} \\
        8 & FLAN Alpaca XL & 3B & \cite{chung2022scaling, wang2022self} & \url{https://huggingface.co/declare-lab/flan-alpaca-xl} \\
        9 & ChatGLM & 6B & \cite{zeng2023glm-130b, du2022glm} & \url{https://huggingface.co/THUDM/chatglm-6b} \\
        10 & ChatGPT & 175B & - & \url{https://platform.openai.com/docs/models/gpt-3-5} \\
        \bottomrule
    \end{tabular}}
    \caption{Instruction-tuned LLMs used in Appendix \ref{sec:method-impl}.}
    \label{tab:llm-details}
\end{table*}

We utilize Open Images v7 as our image captioning dataset $\mathcal{D}_{cap}$ with respect to the provided splits.
For automatic validation with NLI, we use the RoBERTa model fine-tuned on various NLI corpora that achieves the best performance on the Adversarial NLI benchmark~\cite{nie-etal-2020-adversarial}.\footnote{\url{https://huggingface.co/ynie/roberta-large-snli_mnli_fever_anli_R1_R2_R3-nli}} For automatic validation with image-QA pair classification, we build a simple CLIP-based~\cite{radford2021learning} binary classifier. We provide the details in Appendix~\ref{sec:img-txt-cls}. For the \textbf{list-then-rewrite} method, we use $m = 10$.

\subsection{Image-QA Pair Classification}
\label{sec:img-txt-cls}

To construct a model for our image-QA pair classification, we construct a balanced image-QA corpus using \negp~and \others~VQA data randomly selected from 9 existing VQA datasets, i.e., VQAv2 (Balanced Real)~\cite{antol2015vqa},
AdVQA~\cite{sheng2021human}, VizWiz~\cite{gurari2018vizwiz, gurari2019vizwiz}, TextVQA~\cite{Singh_2019_CVPR}, R-VQA~\cite{lu2018rvqa}, Visual7W~\cite{zhu2016visual7w}, TDIUC~\cite{kafle2017analysis}, VQA-Rephrasings~\cite{shah2019cycle}, and VQAv1 (Abstract Scenes)~\cite{antol2015vqa}.

For the image-QA pairs from the \negp~VQA data, we assign a binary label of 1 (valid), which means that the QAs correctly fit the corresponding images as valid pairs. For the \others~VQA data, we replace the \others~ground truth answers with \negp~answers $\in A^{\mathrm{NegP}}$ to make the invalid image-QA pairs (a binary label of 0). We split the corpus into 6k training, 2k validation, and 2k test set.

Using this corpus, we train a simple classifier with one hidden layer on top of a frozen CLIP~\cite{radford2021learning}. We leverage the image-text alignment learned by CLIP~\cite{radford2021learning}, which has been pre-trained on 400M image-text pairs using contrastive learning, to extract the image features of the images and the textual features of their question-answer counterparts. We simply concatenate both image and text features, then input them into the classifier. Our image-QA pair classifier yields an F1-score of 91.29\% on the test set.

\section{Human Evaluation Category Examples}
\label{sec:human-eval-category-ex}

\begin{figure*}[ht]
  \centering
  \includegraphics[width=1.0\linewidth]{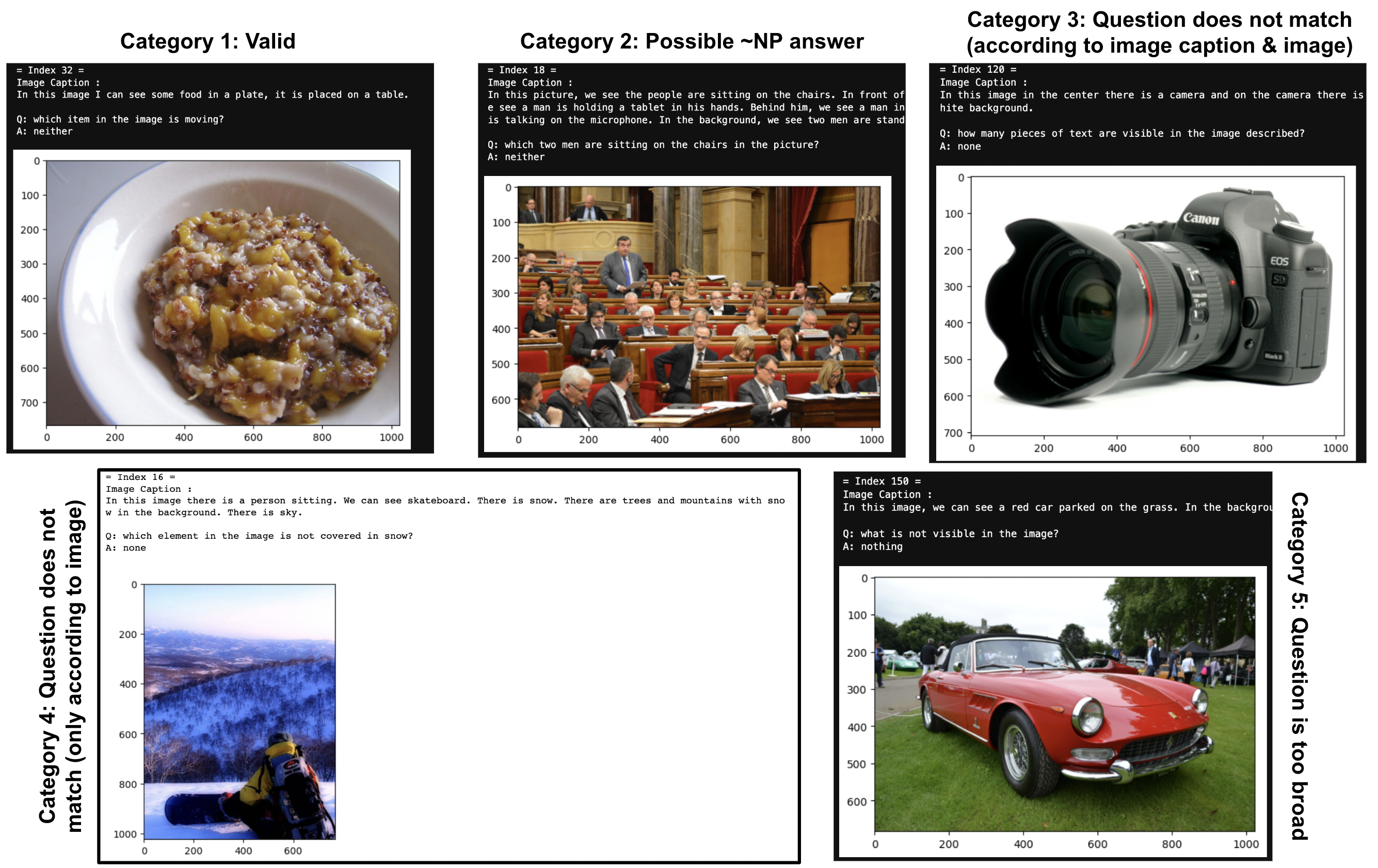}
  \caption{Examples of the human evaluation judgments for the \textbf{generate-from-scratch} prompting method in \S\ref{sec:human-eval-guidelines}.}
  \label{fig:generate-from-scratch-human-eval-categories}
\end{figure*}

We provide the human evaluation categories (\S\ref{sec:human-eval-guidelines}) in Figure~\ref{fig:generate-from-scratch-human-eval-categories}.

\section{Automatic Validation Results of \negp~VQA Data Generation}
\label{sec:auto-validation-results}

\begin{figure}[ht]
  \centering
  \includegraphics[width=0.6\linewidth]{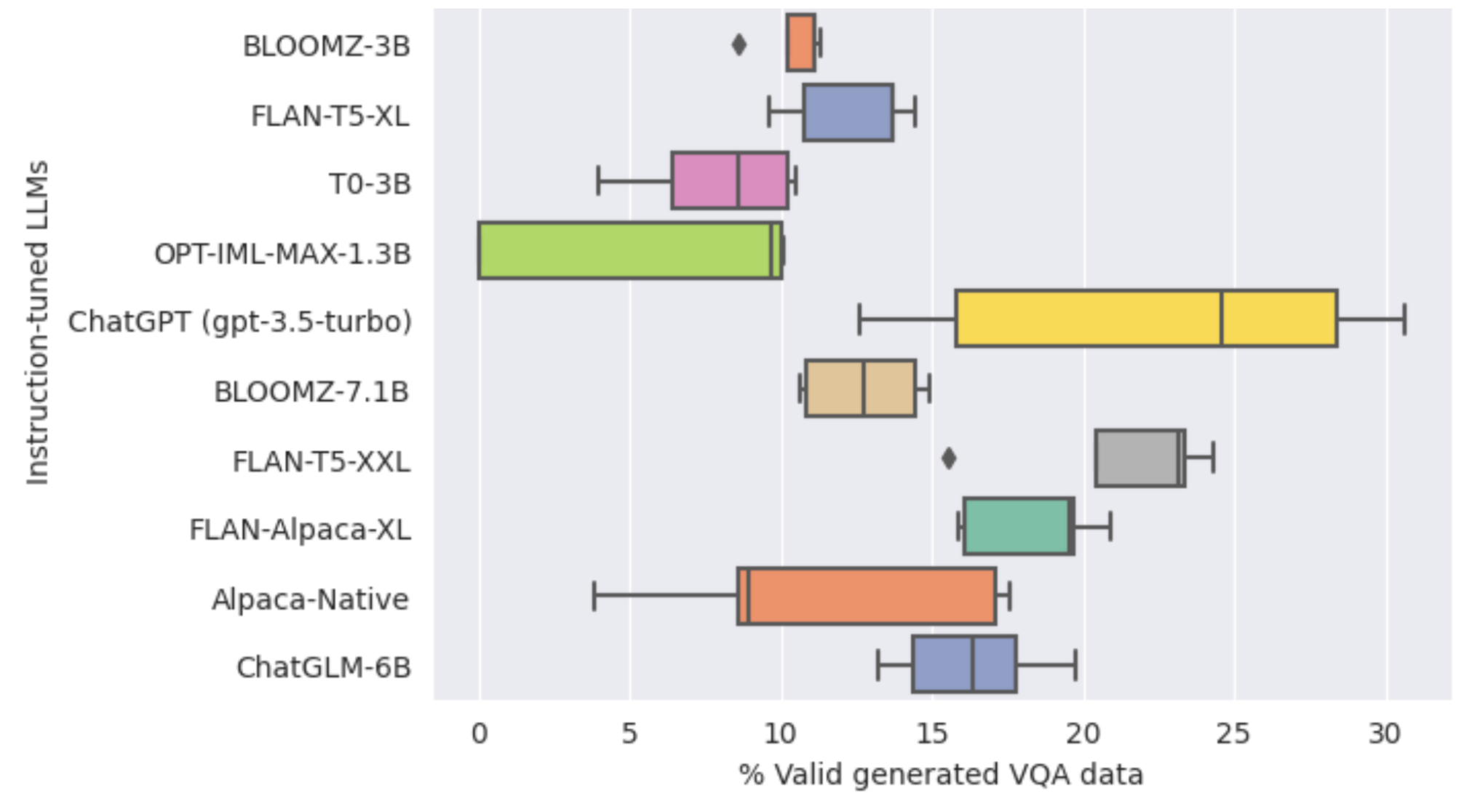}
  \caption{Automatic validation results on 1000 \negp~questions generated using \textbf{generate-from-scratch} (\S\ref{sec:prompt-method}) over five prompt templates.}
  \label{fig:generate-from-scratch-auto-eval}
\end{figure}

\begin{figure}[ht]
  \centering
  \includegraphics[width=0.6\linewidth]{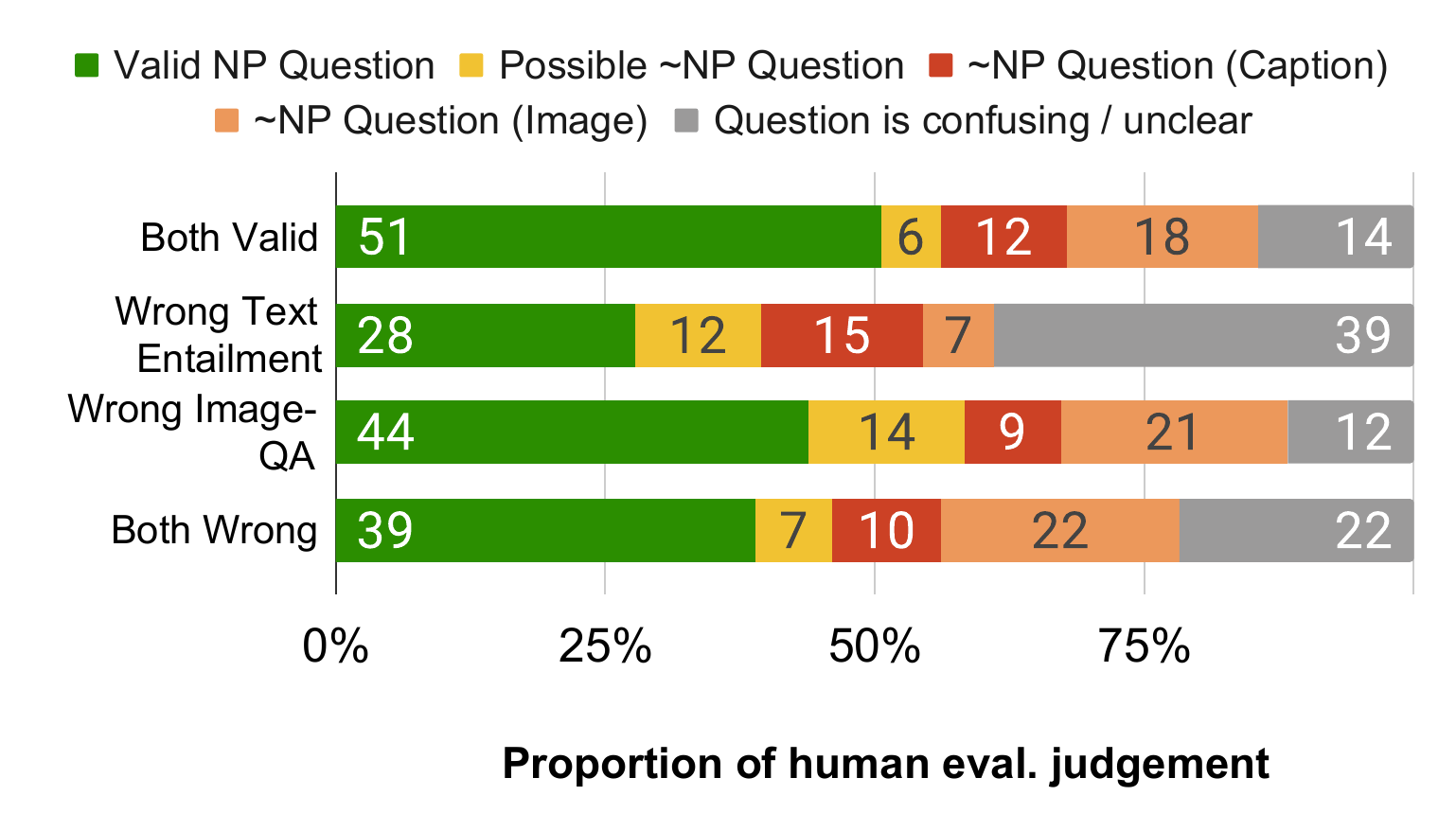}
  \caption{Human evaluation results on \negp~questions generated by ChatGPT using \textbf{generate-from-scratch} (\S\ref{sec:prompt-method}). The Y-axis denotes the verdict from the automatic validators, i.e., caption-QA and image-QA entailment models.}
  \label{fig:generate-from-scratch-human-eval}
\end{figure}

\paragraph{Generate-from-scratch}

Figure~\ref{fig:generate-from-scratch-auto-eval} shows the proportions of valid generated \negp~VQA data using 10 instruction-tuned LLMs listed in Appendix~\ref{sec:method-impl} over five different prompt templates, where each model generates 1k questions per template. The prompt templates are provided in Appendix~\ref{sec:prompt-templates}. The result shows that only $\sim$25\% of the generated questions by the best-performing model, ChatGPT, are valid according to the automatic validation, while other models' valid generated questions range from 6\%-23\%. This indicates that the task of \negp~question generation is more complex and difficult than the instructions used to fine-tune the LLMs.

Next, we conduct a human evaluation on randomly selected 240 generated questions (i.e., 60 for each category in \S\ref{sec:human-eval-guidelines}) by ChatGPT, which is the best-performing model. We ask 3 human experts to judge each generated question and answer pair into one of the five options defined in \S\ref{sec:human-eval-guidelines}. Figure~\ref{fig:generate-from-scratch-human-eval} demonstrates the result of our human evaluation. The result shows that automatic validation judgments do not agree with the human judgments on a considerable amount of the data, even for simple valid/invalid classification, the automatic validation judgments misclassify 27\%-50\% of the subsets. From this result, we can conjecture that our automatic validation approach is not effective at verifying whether the generated \negp~questions are valid or invalid and that the generate-from-scratch prompting method is not reliable and fails to elicit the LLMs' understanding of the task.

\paragraph{List-then-rewrite}

\begin{table}[ht]
    \centering
    \resizebox{0.8\linewidth}{!}{
    \begin{tabular}{p{5cm} p{3cm} p{5cm}}
        \toprule
        \textbf{Instruction-tuned LLM} & \textbf{\% Valid objects} & \textbf{\% Valid objects \& questions} \\
        \midrule
        FLAN T5 XL & 11 & 10 \\
        FLAN T5 XXL & 5 & 17 \\
        Alpaca & 44 & 53 \\
        FLAN Alpaca XL & 25 & 11 \\
        ChatGLM & 84 & 44 \\
        ChatGPT & 99 & \textbf{98} \\
        \bottomrule
    \end{tabular}}
    \caption{Automatic validation results on 100 \negp~questions generated using \textbf{list-then-rewrite} (\S\ref{sec:prompt-method}).}
    \label{tab:list-then-rewrite-auto-eval}
\end{table}

\begin{figure}[ht]
  \centering
  \includegraphics[width=0.6\linewidth]{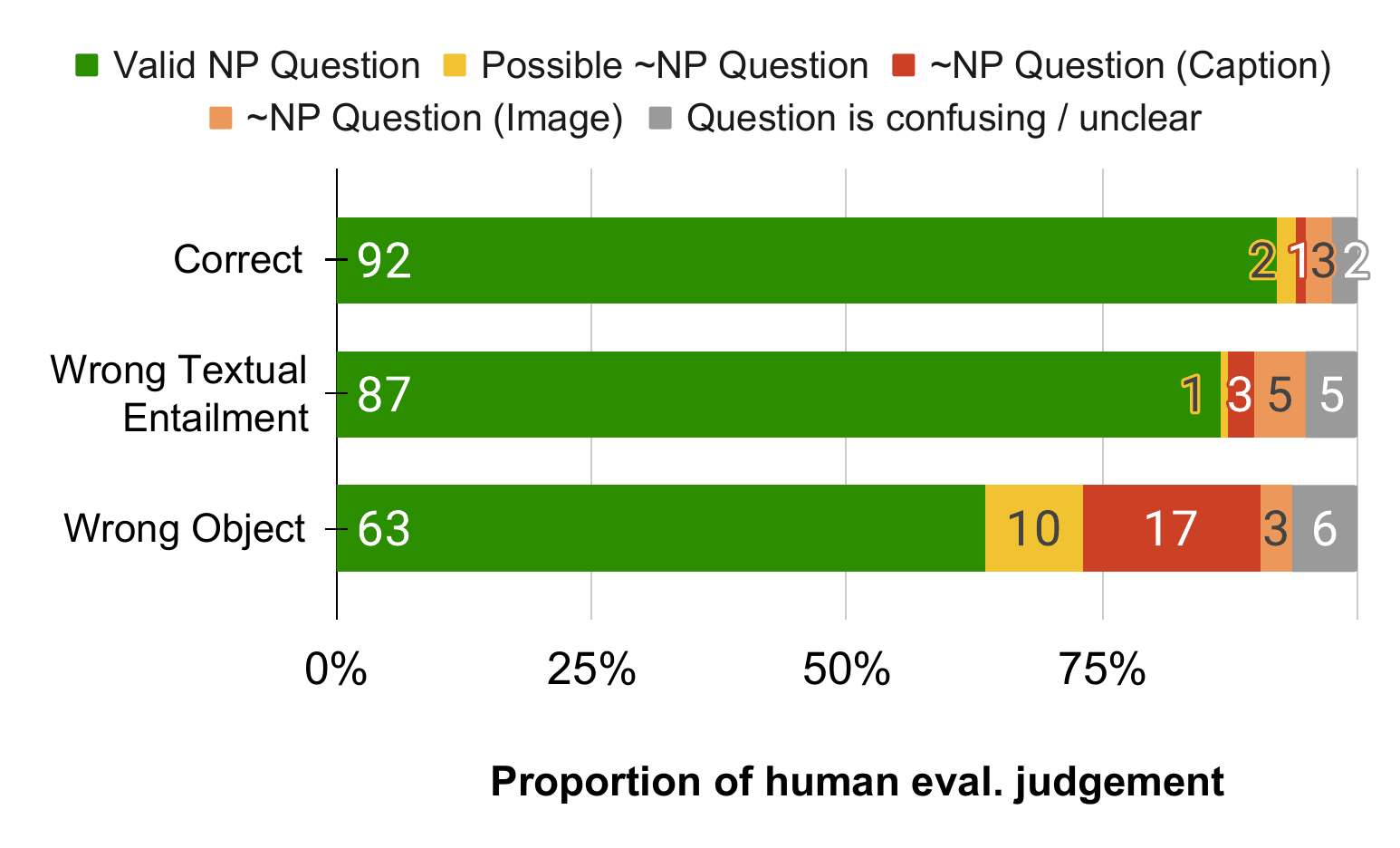}
  \caption{Human evaluation results on \negp~questions generated by ChatGPT using \textbf{list-then-rewrite} (\S\ref{sec:prompt-method}).}
  \label{fig:list-then-rewrite-human-eval}
\end{figure}

The automatic validation results on 100 generated questions (i.e., with the category proportion of 50, 35, and 15, respectively) by \textbf{list-then-rewrite} are provided in Table~\ref{tab:list-then-rewrite-auto-eval}. The best-performing model, ChatGPT, yields 98\% valid questions with a valid non-existent object according to the automatic validation judgments, which is a huge improvement compared to \textbf{generate-from-scratch}. Similarly, Alpaca and ChatGLM also experience the same increase in validity (albeit not as significant), while the FLAN family models deteriorate due to their inability to handle lists inside the instructions, thus forcing them to respond with only one object instead of 10 objects (\S\ref{sec:method-impl}).

Our human evaluation on 300 generated questions by ChatGPT (presented in Figure~\ref{fig:list-then-rewrite-human-eval}) also proves that, when we omit the question generation on the wrong object, we can achieve around 90\% high-quality \negp~questions generated by the \textbf{list-the-rewrite} method. However, this method would benefit from the establishment of a more suitable penalizing method to filter out the generated questions that are inconsistent with the image captions.

\clearpage

\section{Question Diversity of Existing VQA Datasets}
\label{sec:existing-question-diversity}

\begin{figure}[ht]
    \begin{subfigure}[b]{0.5\textwidth}
         \centering
         \includegraphics[width=\textwidth]{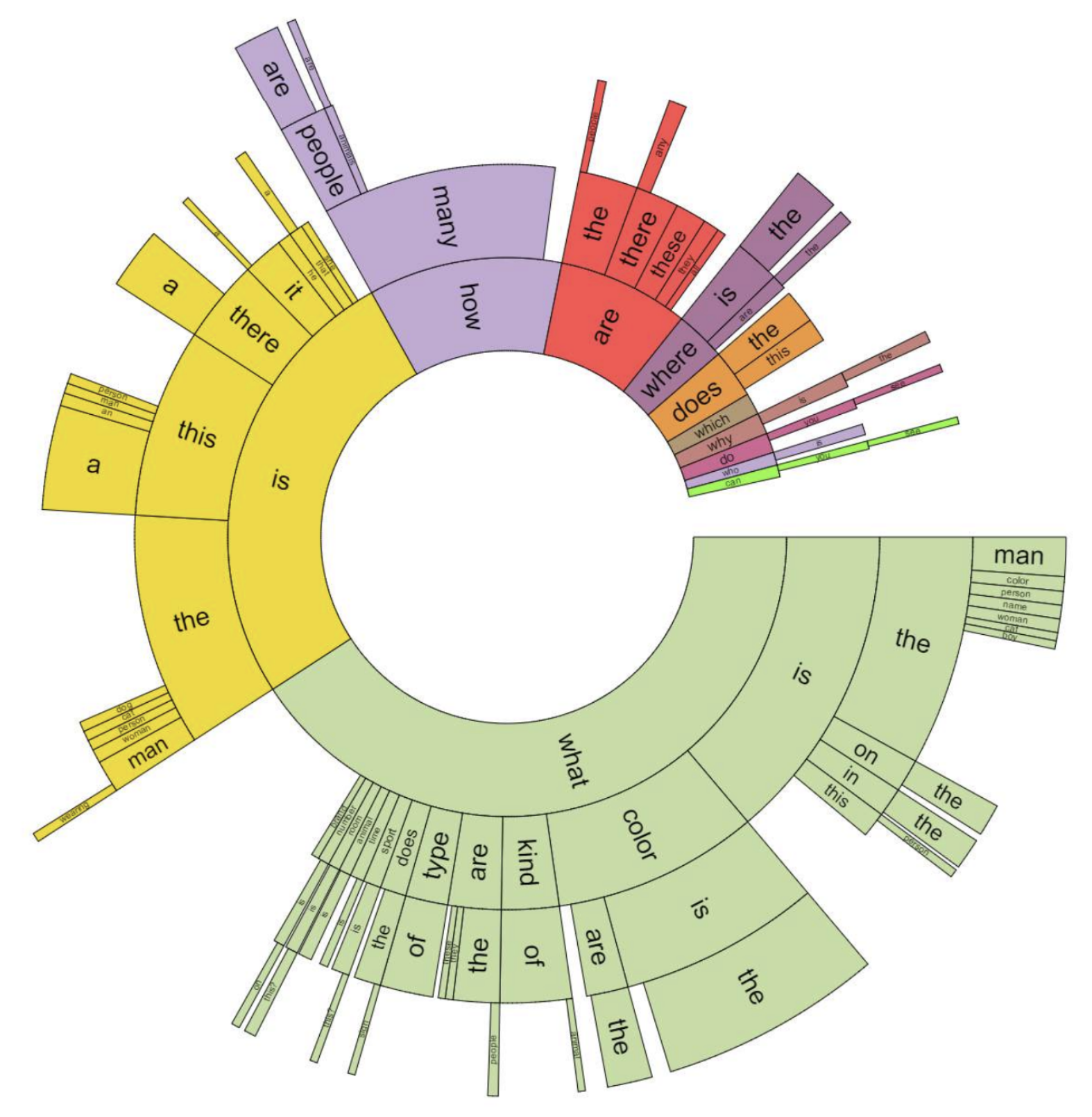}
         \caption{VQAv2 dataset~\cite{antol2015vqa}}
         \label{fig:vqav2-sunburst}
     \end{subfigure}
     \hfill
     \begin{subfigure}[b]{0.5\textwidth}
         \centering
         \includegraphics[width=\textwidth]{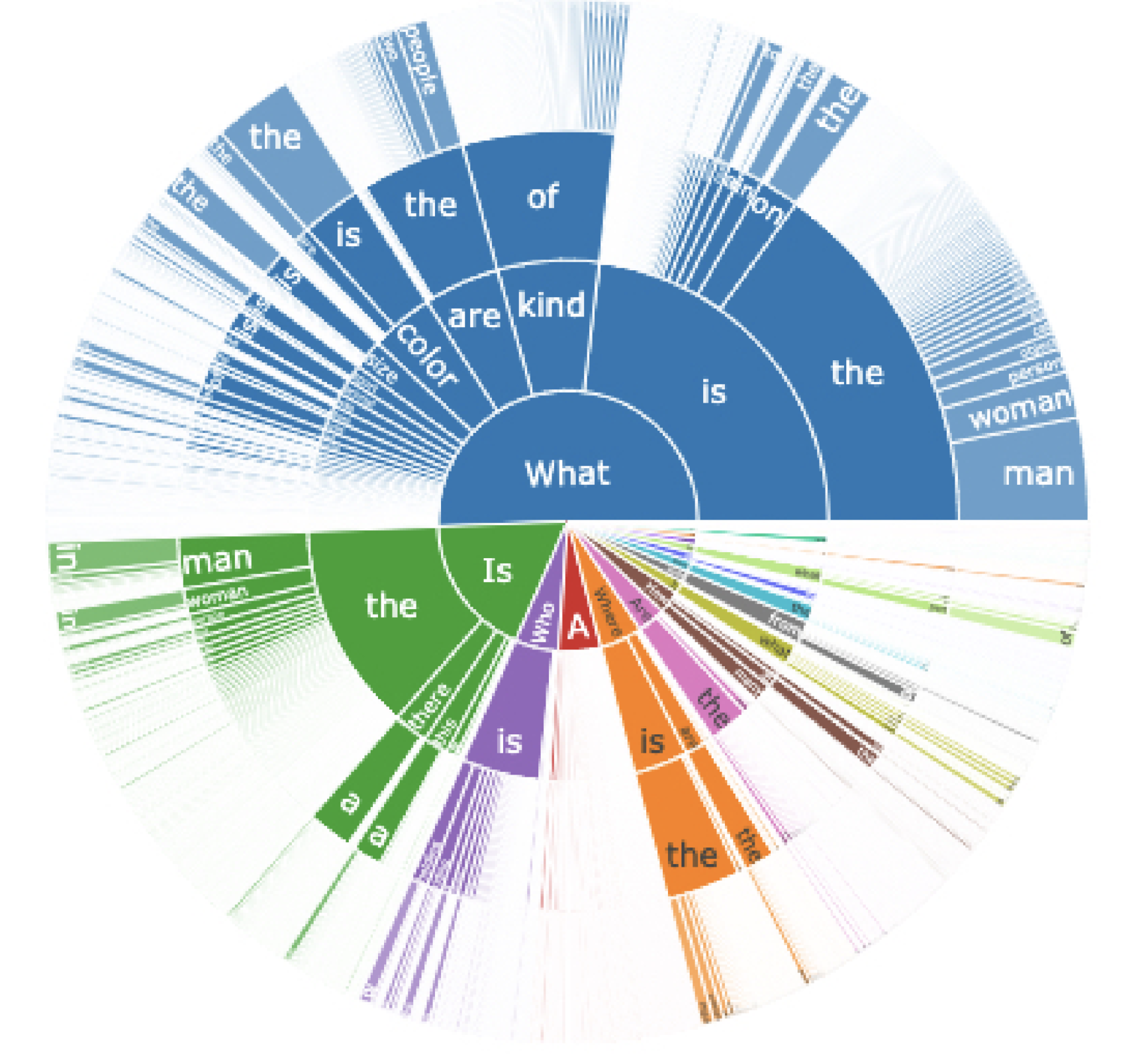}
         \caption{VQ$^2$A dataset~\cite{changpinyo2022all}}
         \label{fig:vq2a-sunburst}
     \end{subfigure}
    \caption{Question diversity of existing datasets. The figures are taken from the respective original papers.}
    \label{fig:prior-works-sunburst}
\end{figure}

We provide the illustrations of question diversity of existing VQA datasets: VQAv2 dataset~\cite{antol2015vqa} which utilizes a manual data generation method (presented in Figure~\ref{fig:vqav2-sunburst}) and VQA-Rephrasings dataset~\cite{shah2019cycle} which utilizes an automatic data generation method (presented in Figure~\ref{fig:vq2a-sunburst}).

\section{Baselines in NOPE Benchmark}
\label{sec:baseline-details}

The variant details of the baselines used in NOPE are presented in Table~\ref{tab:baseline-details}.

\begin{table*}[hb]
    \centering
    \resizebox{1.0\linewidth}{!}{
    \begin{tabular}{l l p{3.5cm} p{12cm}} %p{4cm} p{8cm}}
        \toprule
        \textbf{No} & \textbf{Model} & \textbf{References} & \textbf{Access} \\
        \toprule
        \multicolumn{4}{>{\columncolor[gray]{.9}}c}{\textbf{\textit{Zero-shot \& Few-shot}}} \\
        1 & PromptCap$_{BASE}$ & \cite{hu2022promptcap} & \url{https://huggingface.co/tifa-benchmark/promptcap-coco-vqa}, \url{https://huggingface.co/allenai/unifiedqa-t5-base} \\
        2 & PromptCap & \cite{hu2022promptcap} & \url{https://huggingface.co/tifa-benchmark/promptcap-coco-vqa}, \url{https://huggingface.co/allenai/unifiedqa-t5-3b} \\
        3 & BLIP-2 & \cite{li2023blip} & \url{https://huggingface.co/Salesforce/blip2-opt-2.7b} \\
        4 & OpenFlamingo & \cite{alayrac2022Flamingo, anas_awadalla_2023_7733589} & \url{https://huggingface.co/OpenFlamingo/OpenFlamingo-9B} \\
        5 & InstructBLIP & \cite{instructblip} & \url{https://huggingface.co/Salesforce/instructblip-flan-t5-xl} \\
        \midrule
        \multicolumn{4}{>{\columncolor[gray]{.9}}c}{\textbf{\textit{VQA fine-tuned}}} \\
        1 & OFA & \cite{wang2022ofa} & \url{https://huggingface.co/OFA-Sys/ofa-huge-vqa} \\
        2 & BLIP & \cite{li2022blip} & \url{https://huggingface.co/Salesforce/blip-vqa-base} \\
        3 & BLIP$_{CapFilt-L}$ & \cite{li2022blip} & \url{https://huggingface.co/Salesforce/blip-vqa-capfilt-large} \\
        4 & ALBEF & \cite{li2021align} & \url{https://github.com/salesforce/ALBEF\#download\#Finetuned-checkpoint-for-VQA} \\
        5 & GIT$_{LARGE}$ & \cite{wang2022git} & \url{https://huggingface.co/microsoft/git-large-vqav2} \\
        \bottomrule
    \end{tabular}}
    \caption{Variant details of the baselines in NOPE (\S\ref{sec:nope-baselines}).}
    \label{tab:baseline-details}
\end{table*}

% \clearpage

\section{Examples of Object Hallucination in NOPE}
\label{sec:obj-hall-examples}

We list the examples of object hallucination from the dev set of NOPE in Table~\ref{tab:obj-hall-examples}.

\begin{table*}
    \centering
    \resizebox{1.0\linewidth}{!}{
    \begin{tabular}{c p{2.5cm} c c p{9cm}}
        \toprule
        \textbf{ID} & \textbf{Object-scene relevance} & \textbf{Visual context} & \multicolumn{2}{c}{\textbf{Question-answer}} \\
        % \cmidrule(lr){5-7} \cmidrule(lr){8-12}
         % & & & & PromptCap-3B & OpenFlamingo & BLIP-2 & OFA & BLIP & ALBEF & GIT$_{LARGE}$ & InstructBLIP \\
         \toprule
        \multirow{10}{*}{390} & \multirow{10}{*}{Related} & \multirow{10}{*}{\includegraphics[width=0.47\linewidth]{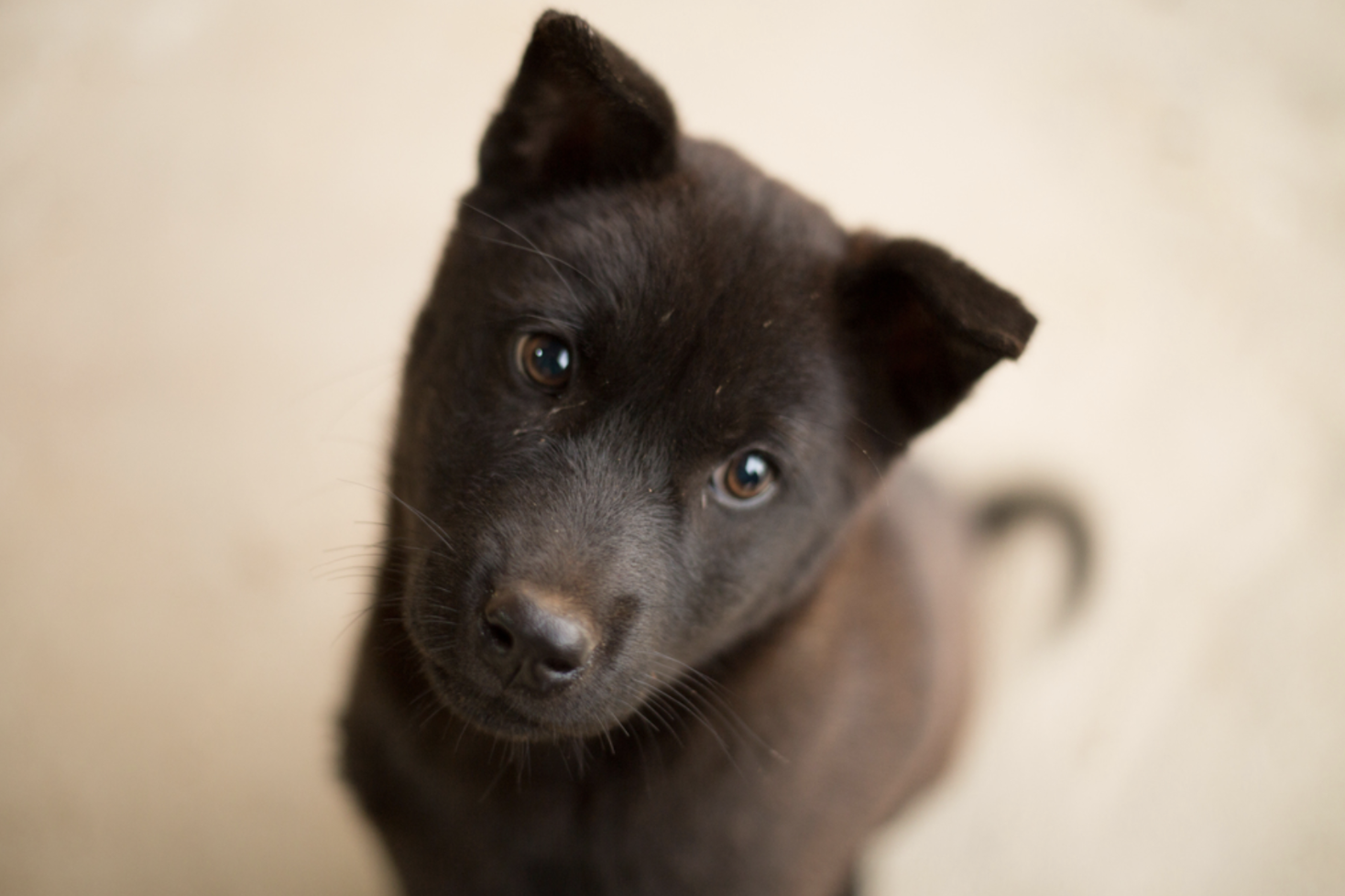}} & \textbf{Question} & can you point out the location of the dog collar in the image? \\
         & & & \textbf{GT answer} & nowhere \\
         \cmidrule{4-5}
         & & & \textbf{PromptCap} & \textcolor{np_color}{yes} \\
         & & & \textbf{OpenFlamingo} & \textcolor{np_color}{The dog collar is located on the dog's neck.} \\
         & & & \textbf{BLIP-2} & \textcolor{np_color}{yes, it is on the dog's collar} \\
         & & & \textbf{OFA} & no \\
         & & & \textbf{BLIP} & no \\
         & & & \textbf{ALBEF} & \textcolor{np_color}{dog's neck} \\
         & & & \textbf{GIT$_{LARGE}$} & no \\
         & & & \textbf{InstructBLIP} & no \\
         \midrule
         \multirow{10}{*}{822} & \multirow{10}{*}{Related} & \multirow{10}{*}{\includegraphics[width=0.23\linewidth]{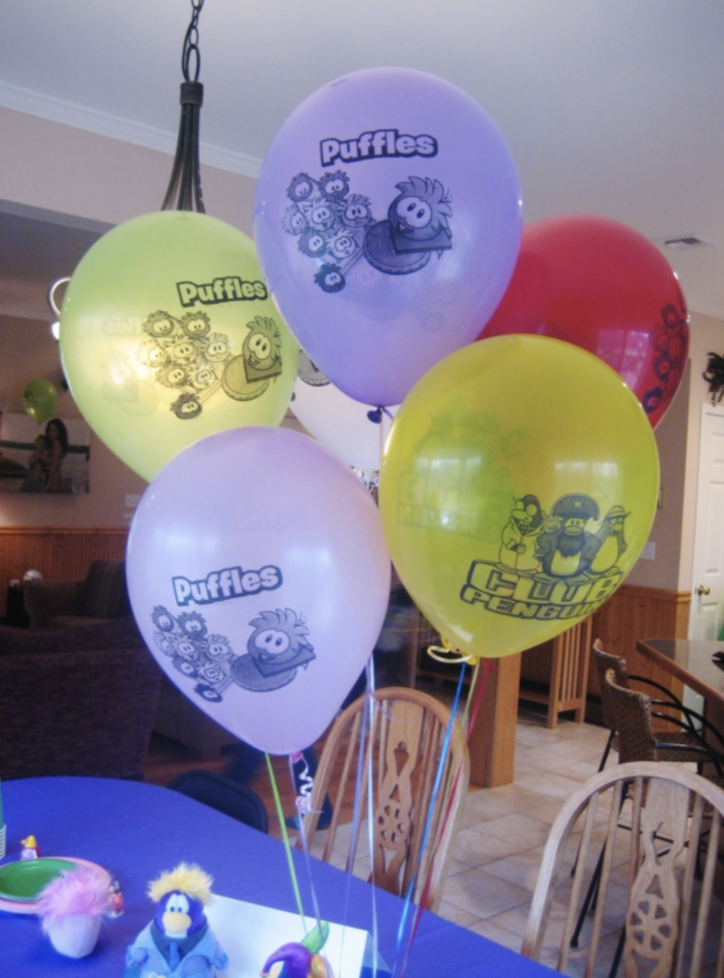}} & \textbf{Question} & how many menu cards have been captured in the image? \\
         & & & \textbf{GT answer} & none \\
         \cmidrule{4-5}
         & & & \textbf{PromptCap} & \textcolor{np_color}{1} \\
         & & & \textbf{OpenFlamingo} & \textcolor{np_color}{10} \\
         & & & \textbf{BLIP-2} & none \\
         & & & \textbf{OFA} & \textcolor{np_color}{8} \\
         & & & \textbf{BLIP} & \textcolor{np_color}{six} \\
         & & & \textbf{ALBEF} & \textcolor{np_color}{2,3} \\
         & & & \textbf{GIT$_{LARGE}$} & 0 \\
         & & & \textbf{InstructBLIP} & 0 \\
         \midrule
         \multirow{10}{*}{982} & \multirow{10}{*}{Related} & \multirow{10}{*}{\includegraphics[width=0.41\linewidth]{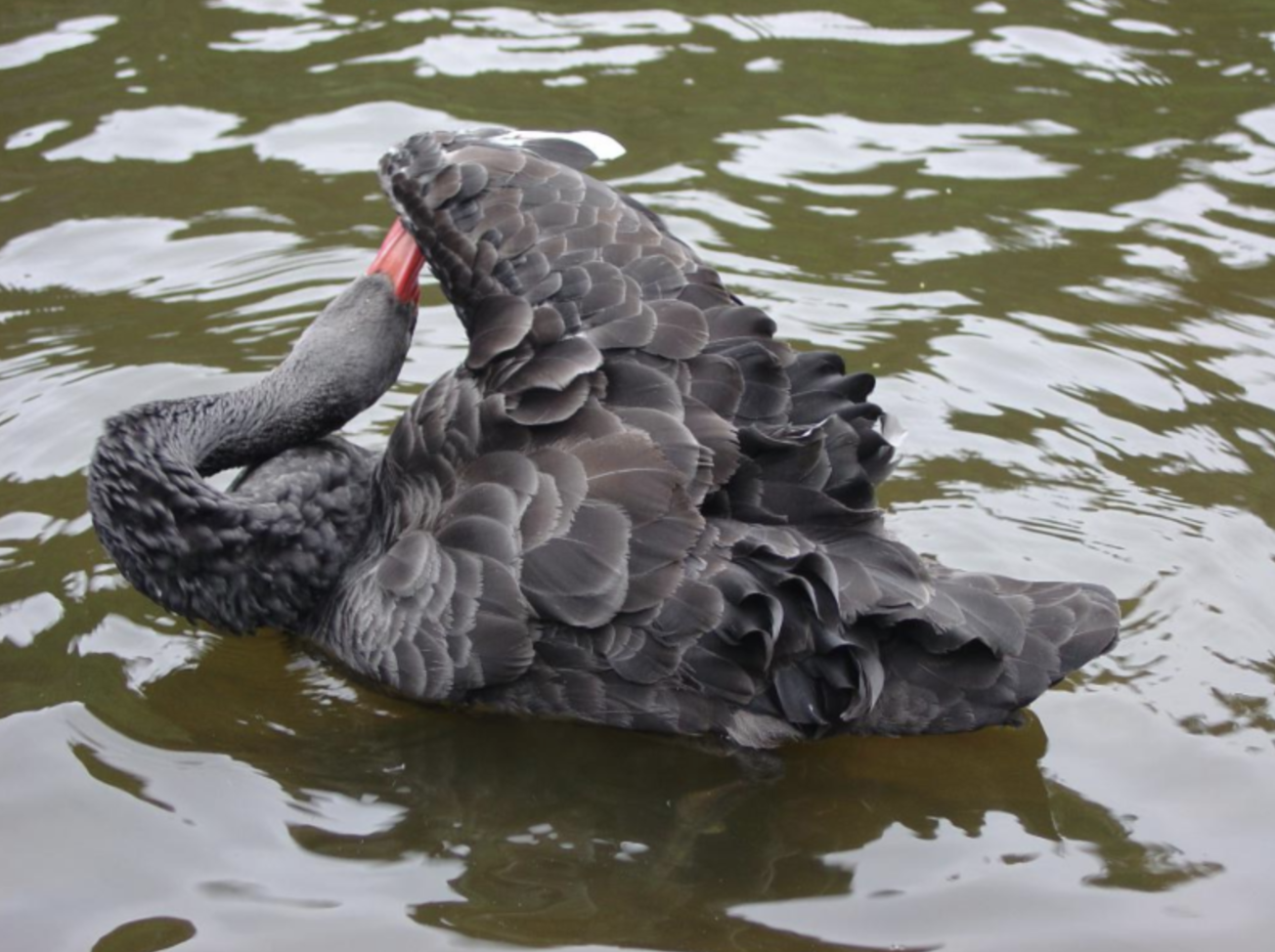}} & \textbf{Question} & where can the clouds be spotted in the image? \\
         & & & \textbf{GT answer} & nowhere \\
         \cmidrule{4-5}
         & & & \textbf{PromptCap} & \textcolor{np_color}{in the sky} \\
         & & & \textbf{OpenFlamingo} & \textcolor{np_color}{On the head of a goose.} \\
         & & & \textbf{BLIP-2} & \textcolor{np_color}{the clouds are in the sky} \\
         & & & \textbf{OFA} & \textcolor{np_color}{the sky} \\
         & & & \textbf{BLIP} & \textcolor{np_color}{yes} \\
         & & & \textbf{ALBEF} & \textcolor{np_color}{in the snow.} \\
         & & & \textbf{GIT$_{LARGE}$} & no \\
         & & & \textbf{InstructBLIP} & \textcolor{np_color}{sky} \\
         \midrule
         \multirow{10}{*}{9165} & \multirow{10}{*}{Partially related} & \multirow{10}{*}{\includegraphics[width=0.41\linewidth]{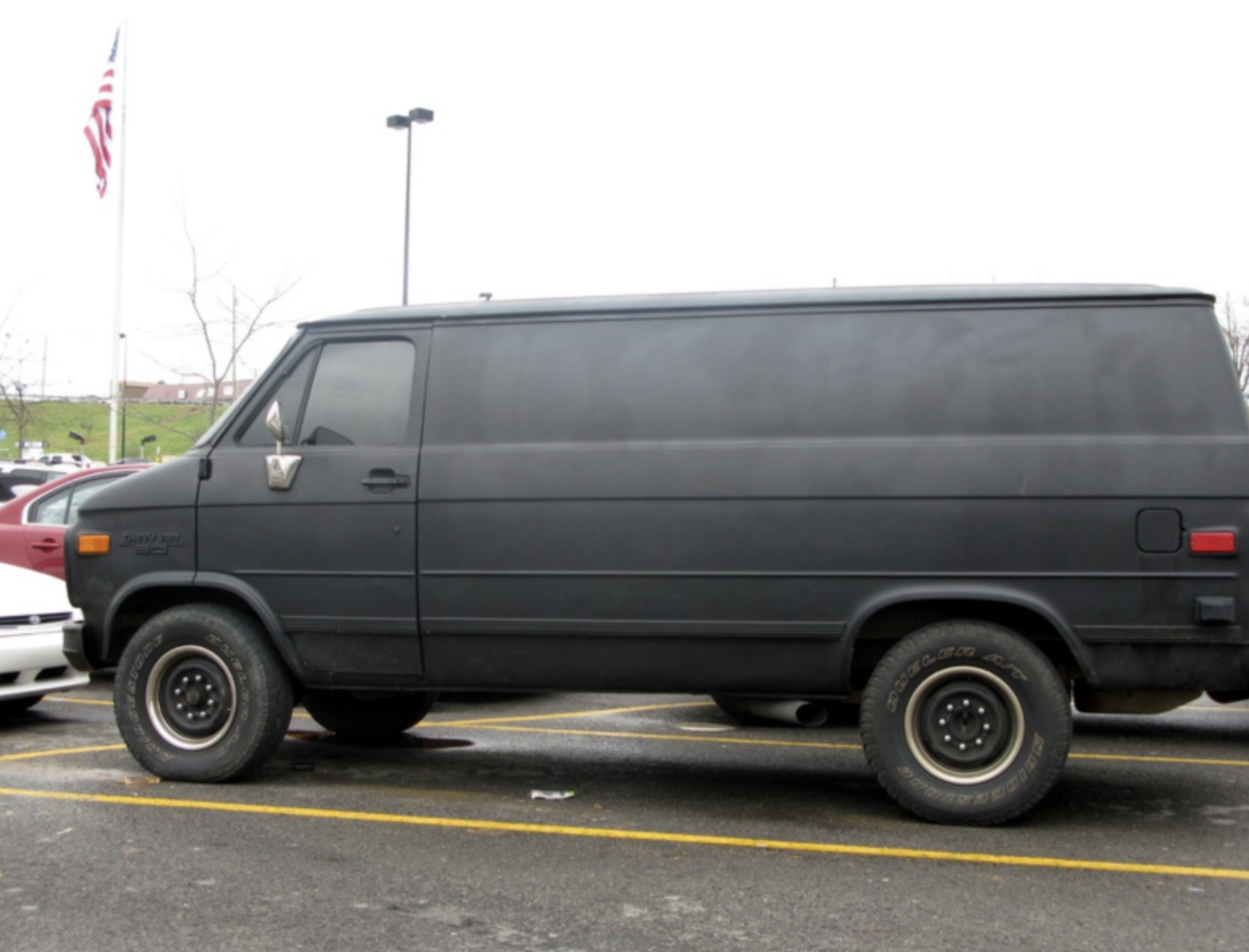}} & \textbf{Question} & who can you see using the fishing rod? \\
         & & & \textbf{GT answer} & nobody \\
         \cmidrule{4-5}
         & & & \textbf{PromptCap} & \textcolor{np_color}{a gray van} \\
         & & & \textbf{OpenFlamingo} & \textcolor{np_color}{The owner of this Dodge B250 van.} \\
         & & & \textbf{BLIP-2} & \textcolor{np_color}{the guy in the back of the van} \\
         & & & \textbf{OFA} & no 1 \\
         & & & \textbf{BLIP} & no idea \\
         & & & \textbf{ALBEF} & \textcolor{np_color}{man dancing} \\
         & & & \textbf{GIT$_{LARGE}$} & no \\
         & & & \textbf{InstructBLIP} & no one \\
         \midrule
         \multirow{10}{*}{10135} & \multirow{10}{*}{Unrelated} & \multirow{10}{*}{\includegraphics[width=0.41\linewidth]{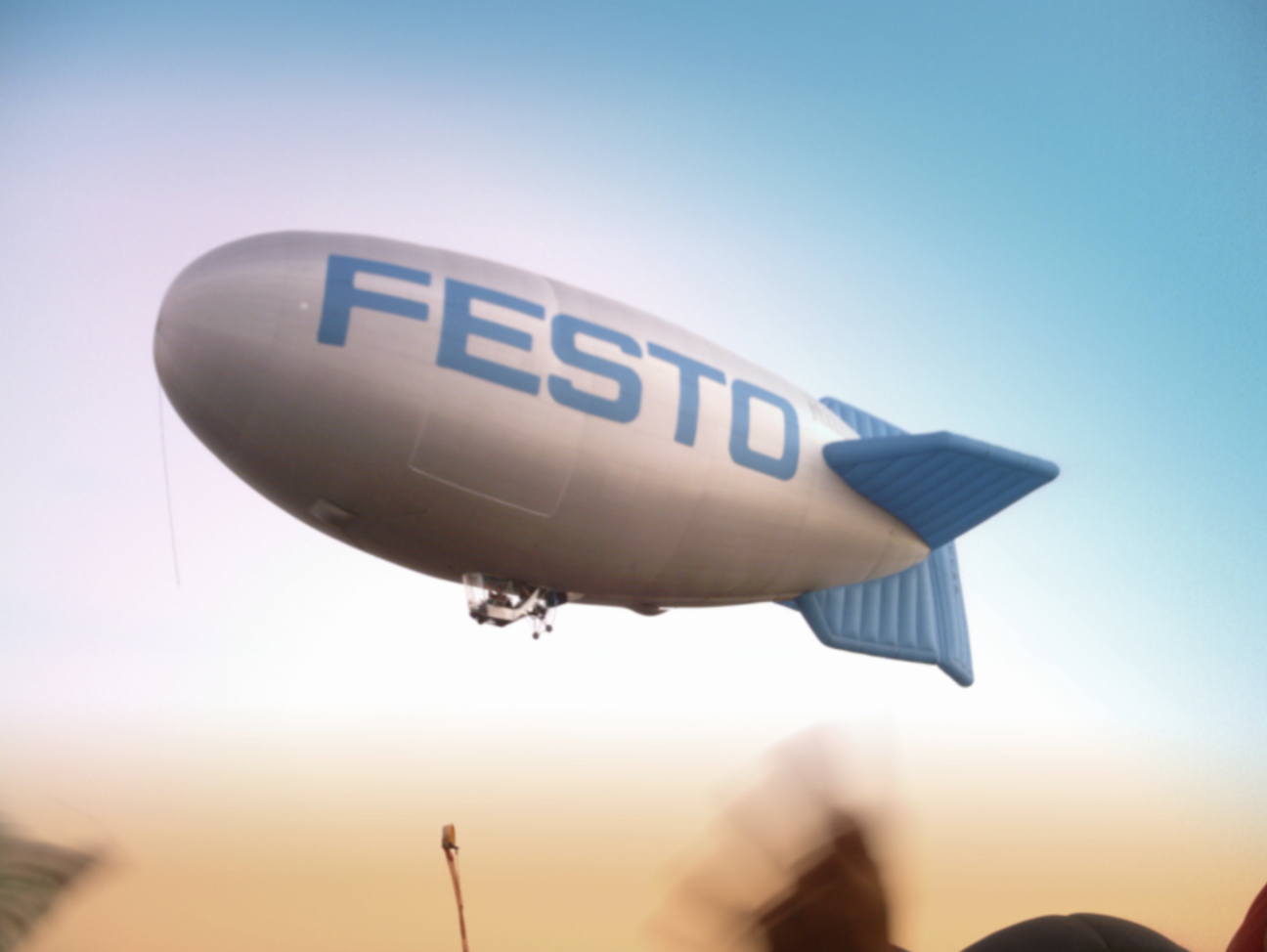}} & \textbf{Question} & which color is the pillow in the image? \\
         & & & \textbf{GT answer} & nothing \\
         \cmidrule{4-5}
         & & & \textbf{PromptCap} & \textcolor{np_color}{blue} \\
         & & & \textbf{OpenFlamingo} & \textcolor{np_color}{blue} \\
         & & & \textbf{BLIP-2} & \textcolor{np_color}{blue} \\
         & & & \textbf{OFA} & \textcolor{np_color}{black} \\
         & & & \textbf{BLIP} & \textcolor{np_color}{red and white} \\
         & & & \textbf{ALBEF} & \textcolor{np_color}{red black white} \\
         & & & \textbf{GIT$_{LARGE}$} & \textcolor{np_color}{blue} \\
         & & & \textbf{InstructBLIP} & \textcolor{np_color}{white} \\
         \bottomrule
    \end{tabular}}
    \caption{Examples of object hallucination in the dev set of NOPE. The hallucinated answers are shown in \textcolor{np_color}{\textbf{pink}}.}
    \label{tab:obj-hall-examples}
\end{table*}

\end{document}